\documentclass{article} 
\usepackage{iclr2027_conference,times}

\usepackage{amsmath,amsfonts,bm}

\def\eqref#1{equation~\ref{#1}}

\def\1{\bm{1}}

\DeclareMathAlphabet{\mathsfit}{\encodingdefault}{\sfdefault}{m}{sl}
\SetMathAlphabet{\mathsfit}{bold}{\encodingdefault}{\sfdefault}{bx}{n}

\usepackage{hyperref}
\usepackage{url}
\usepackage[table]{xcolor}
\usepackage{amsmath,amssymb}
\usepackage{graphicx}
\usepackage{subcaption}
\usepackage{booktabs}
\usepackage{tabularx}
\usepackage{float}
\usepackage{adjustbox}
\usepackage{multirow}
\usepackage{makecell}
\usepackage{array}
\usepackage{longtable}
\usepackage{calc} 
\usepackage{etoolbox}
\usepackage{wrapfig}
\usepackage{enumitem}
\usepackage[most]{tcolorbox}

\title{Clarify the User or Verify the World?\\ Uncertainty Routing for Proactive Agents}

\author{
Zhaofeng Li$^{1}$ \qquad
Xuan Zhang$^{1}$ \qquad
Xiaokui Xiao$^{1}$ \qquad
Yang Deng$^{2}$ \\[4pt]
$^{1}$National University of Singapore
\qquad
$^{2}$Singapore Management University \\[3pt]
\texttt{liz88@u.nus.edu} \qquad
\texttt{xuanzhang@u.nus.edu} \\
\texttt{xkxiao@nus.edu.sg} \qquad
\texttt{ydeng@smu.edu.sg}
}

\iclrfinalcopy 
\begin{document}

\maketitle
\lhead{}

\begin{abstract}
Tool-using LLM agents must decide not only whether additional information is needed, but also which source can resolve the uncertainty. Existing proactive approaches often specialize in either user clarification or environment verification, without explicitly determining the appropriate information source for each decision. We formulate this problem as \emph{uncertainty routing} among \textsc{act}, \textsc{clarify}, and \textsc{verify}, and propose \textsc{Prour}, a proactive uncertainty routing framework. \textsc{Prour} decomposes action uncertainty into disagreement across plausible user-goal interpretations, which signals user-side ambiguity, and the entropy remaining within each interpretation, which signals missing world-side evidence. To acquire information from the routed source, a query generator is trained with a mode-conditioned information-gain reward, targeting user-goal identification under \textsc{clarify} and next-action identification under \textsc{verify}. On $\tau$-bench, \textsc{Prour} achieves 28.17\% average success rate across retail and airline, outperforming the strongest prior method by 4.57\% while using 2.17 fewer interaction steps. The learned policy further generalizes to stronger task agents and transactional domains of $\tau^3$-bench without retraining, demonstrating the benefit of source-aligned uncertainty resolution for proactive agents.\footnote{Code is available at
\url{https://anonymous.4open.science/r/Prour-FBC3/}.}
\end{abstract}

\section{Introduction}

Tool-using LLM agents increasingly interact with users, query databases, invoke APIs, and modify external state \citep{yao2022react,schick2023toolformer,yao2024taubench}. To ensure reliability, an agent must recognize when the available information is insufficient and proactively acquire missing information before acting \citep{zhang2024askbeforeplan}. 
The challenge, however, is not merely deciding \emph{whether} to seek more information. A proactive agent must also decide \emph{where} that information should come from. Should it ask the \textbf{user}, or should it inspect the external \textbf{world}?

\begin{figure}[h]
\vspace{-1mm}
    \centering
    \includegraphics[width=\textwidth]{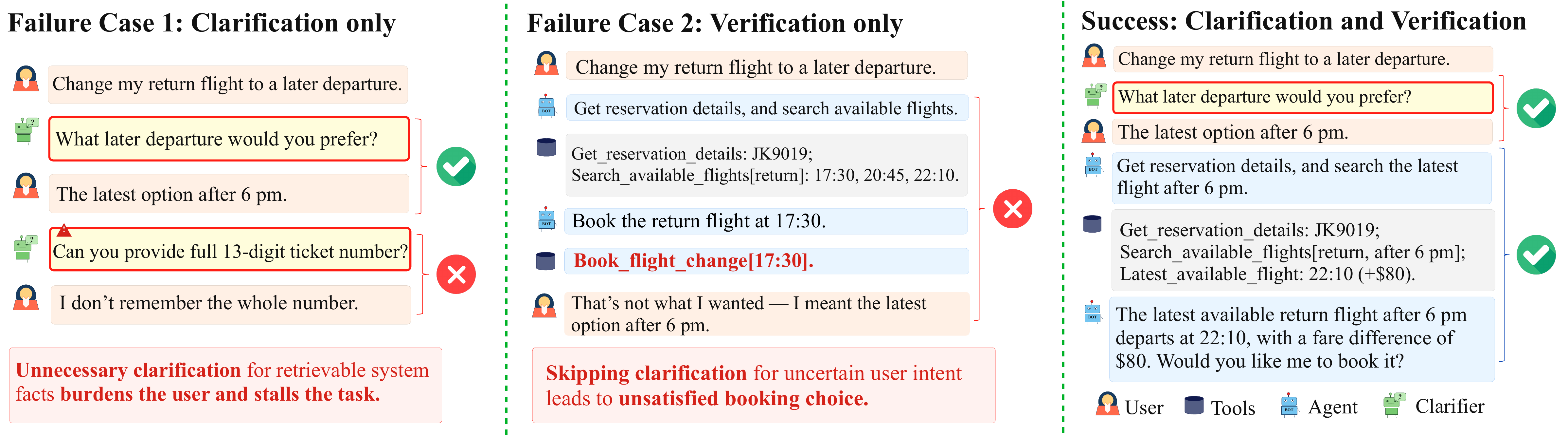}
    \caption{An example of clarification and verification in $\tau$-airline trajectories.
    Clarification alone burdens the user with facts
    available from tools (left), while verification alone misses user preferences (middle). Combining them
    resolves the preference before retrieving system evidence (right).}
    \label{fig:tau_airline_example}
    \vspace{-1mm}
\end{figure}

\begin{wrapfigure}{r}{0.55\linewidth}
\setlength{\abovecaptionskip}{2pt}
\setlength{\belowcaptionskip}{2pt}
    \centering
    \vspace{-5mm}
    \includegraphics[width=\linewidth]{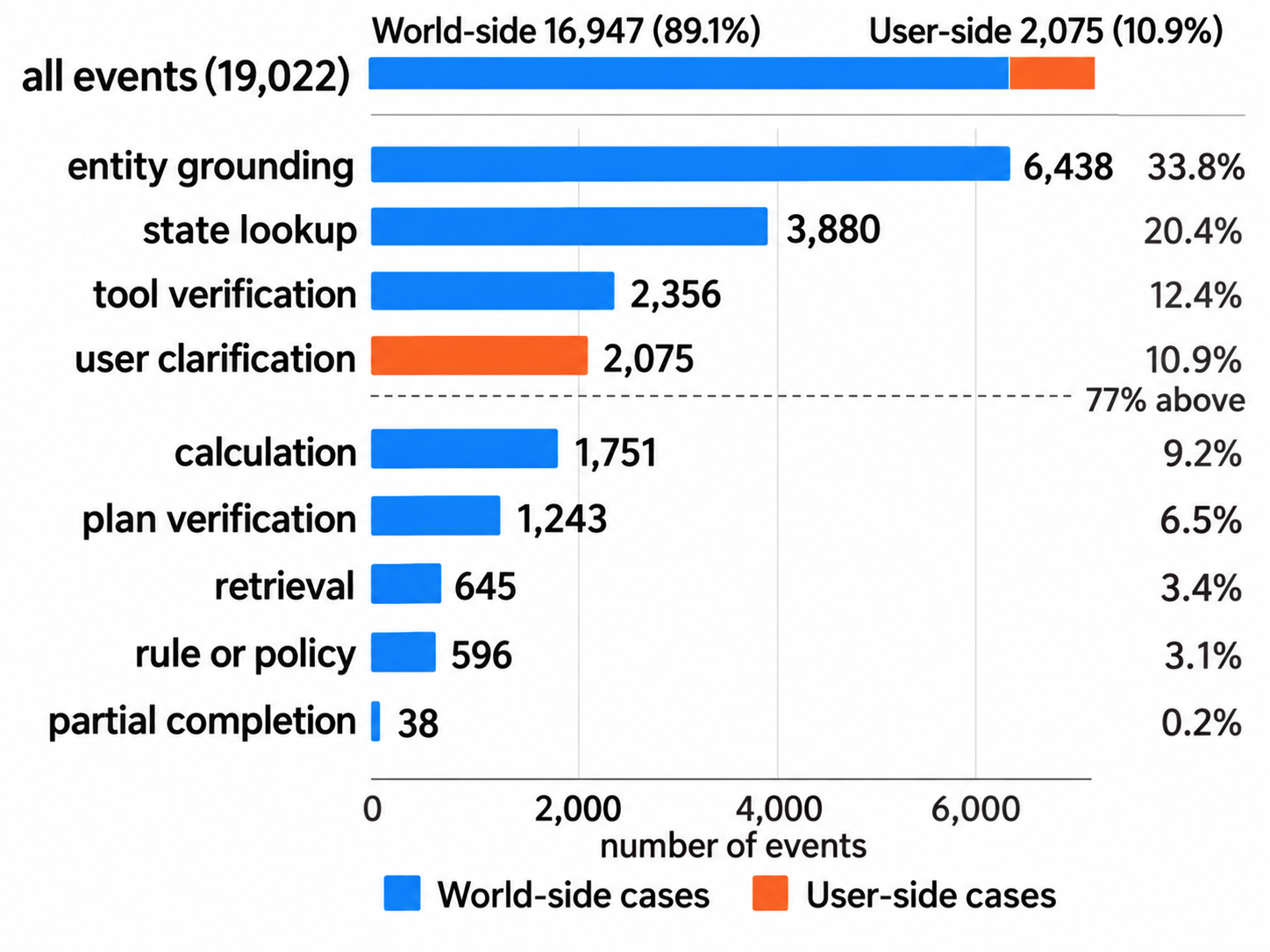}
    \caption{Source distribution of 19,022 step-level events from $\tau$-bench.
    World-side cases (blue) account for 89.1\%; user-side cases (orange) account for 10.9\%. See
    Appendix~\ref{app:uncertainty_case_taxonomy} for the full taxonomy.}
    \label{fig:uncertainty_cases_distribution}
    \vspace{-3mm}
\end{wrapfigure}

This distinction is critical because these two interventions resolve fundamentally different unknowns, as illustrated by the request in Figure~\ref{fig:tau_airline_example}.
In \textbf{Failure Case 1}, clarification effectively recovers the user's preference but then asks the user for a ticket number retrievable from tools, burdening the user and stalling the task. In \textbf{Failure Case 2}, verification retrieves the reservation and available flights but does not clarify what \textit{later} means, leading to a wrong booking. 
Only the \textbf{Success case} obtains information from the appropriate source: it asks the user to resolve the preference and queries the environment to verify eligibility and availability.
These cases expose two distinct information sources: the user side and the world side. Only the user can clarify an unstated preference, whereas world facts such as system state, policy constraints, and action validity should be verified from tools and other grounded evidence. Our analysis of 19,022 step-level events from $\tau$-bench~\citep{yao2024taubench} 
further reveals a strong asymmetry in where missing information resides (Figure~\ref{fig:uncertainty_cases_distribution}): 89.1\% concern world-side information, whereas only 10.9\% concern user-side information. Thus, treating user clarification as the sole mechanism for resolving uncertainty~\citep{deng2026uncertainty,suri2026structured} can be inefficient and often directs the agent to the wrong information source. A proactive agent must determine not only whether it is uncertain, but also whether that uncertainty should be resolved by the user or by the world. We refer to this problem as \emph{uncertainty routing}. 

To address this problem, we propose a \textbf{Pro}active  \textbf{U}ncertainty \textbf{R}outing framework, namely \textbf{\textsc{Prour}}, which explicitly identifies the source of uncertainty before acquiring additional information. \textsc{Prour} decomposes uncertainty over the agent's next action into two source-aligned components, inspired by the distinction between aleatoric and epistemic uncertainty \citep{hou2024decomposing,ling2024uncertainty}.
Specifically, 
if different plausible interpretations of the user's goal imply different actions, the agent faces high \textbf{aleatoric uncertainty (AU)}: the goal itself is underspecified, so the missing information must come from the user. In contrast, if the appropriate action remains uncertain even after ambiguity in the user's goal is clarified, the agent faces high \textbf{epistemic uncertainty (EU)}: it lacks sufficient knowledge about the external state or constraints and should therefore seek additional evidence. This yields a natural three-way routing policy: \textsc{clarify} when AU is high, \textsc{verify} when the goal is clear but EU remains high, and \textsc{act} when neither source of uncertainty warrants further information acquisition.

Routing, however, only determines \emph{where} to look; the agent must still decide \emph{what} to ask. \textsc{Prour} therefore couples uncertainty routing with a mode-conditioned \textbf{Query Generator}. Under \textsc{clarify} mode, the generator issues user-directed queries that make the underlying goal more identifiable; under \textsc{verify} mode, it queries a tool-grounded verifier for evidence that makes the reference next action more identifiable. We train this generator using Decoupled Clip and Dynamic Sampling Policy Optimization (DAPO) \citep{yu2025dapo} with a \textbf{mode-conditioned information-gain reward}. The reward measures how much a query--response pair increases the likelihood of the ground-truth user goal under \textsc{clarify} mode, or of the reference next action under \textsc{verify} mode. Thus, rather than maintaining separate mechanisms for clarifying users and verifying the world, a single policy learns to generate queries whose target is aligned with the source of uncertainty.

Our main contributions are three-fold:
\begin{enumerate}[leftmargin=*,nosep]
    \item 
    We identify uncertainty routing as a fundamental problem for tool-using agents, and propose \textsc{Prour},  a proactive uncertainty routing framework that decomposes action uncertainty into user-side and world-side uncertainty to route the agent among \textsc{act}, \textsc{clarify}, and \textsc{verify}.

    \item 
    We propose a mode-conditioned information-gain objective that aligns both the information source and reward target with the routed uncertainty, and use it to train a single query generator to acquire information from either users or tool-grounded evidence.

    \item Extensive experiments demonstrate the effectiveness, efficiency, and generalization of \textsc{Prour}. On $\tau$-bench, it improves average success rate by 4.57\% with 2.17 fewer interaction steps. Without further retraining, it achieves 29.00\% average success rate on $\tau^3$-bench and generalizes across stronger task agents.

\end{enumerate}

\section{Related Work}
\label{sec:related_work}

\textbf{Uncertainty Decomposition for LLM Agents.}~
For an interactive agent, uncertainty is actionable only if it identifies both the missing information and its source. Classical work separates aleatoric and epistemic uncertainty \citep{kendall2017uncertainties}, and Bayesian Active Learning by Disagreement (BALD) measures parameter-induced disagreement \citep{houlsby2011bald}. LLM studies use calibrated confidence, semantic consistency, or input-clarification ensembles \citep{jiang2021know,kadavath2022language,kuhn2023semantic,hou2024decomposing}. Unlike classical BALD, which
ensembles over model parameters and associates ensemble disagreement with
epistemic uncertainty, we ensemble over candidate goal interpretations. 
Following the input-clarification view of \citet{hou2024decomposing}, disagreement across interpretations captures user-side ambiguity, while within-interpretation entropy captures residual world-side uncertainty. 
We therefore use AU and EU as operational source labels rather than universal Bayesian identities. Prior decompositions diagnose uncertainty but do not route each decision among acting, querying the user, and consulting the world.

\textbf{Proactive Clarification and Verification.}~
Prior work treats clarification as proactive information acquisition, using expected value or predictive uncertainty to decide when to ask \citep{rao2018evpi,naszadi2023predictive}, using agents to clarify before planning \citep{deng2023proactive,zhang2024askbeforeplan}, or modeling latent intents and parameter uncertainty \citep{zhang2025clarify,kobalczyk2025active,suri2026structured}. Learning-based methods optimize long-horizon collaboration, belief gain toward user goals or search answers \citep{wu2025collabllm,deng2026uncertainty,wang2026igpo}. Complementary work uses tool-grounded critique or verification
\citep{gou2024critic,han2025verifiagent} or requires user confirmation \citep{cuadron2025saber}. These approaches typically assume a predefined information source. \textsc{Prour} instead first routes uncertainty to its appropriate source, then uses a shared policy whose answer source and reward target adapt to \textsc{clarify} or \textsc{verify}.

\section{Methodology}
\label{sec:method}

\begin{figure}[t]
    \setlength{\abovecaptionskip}{3pt}
    \setlength{\belowcaptionskip}{2pt}
    \centering
    \begin{subfigure}[t]{0.66\textwidth}
        \centering
        \includegraphics[width=\linewidth]{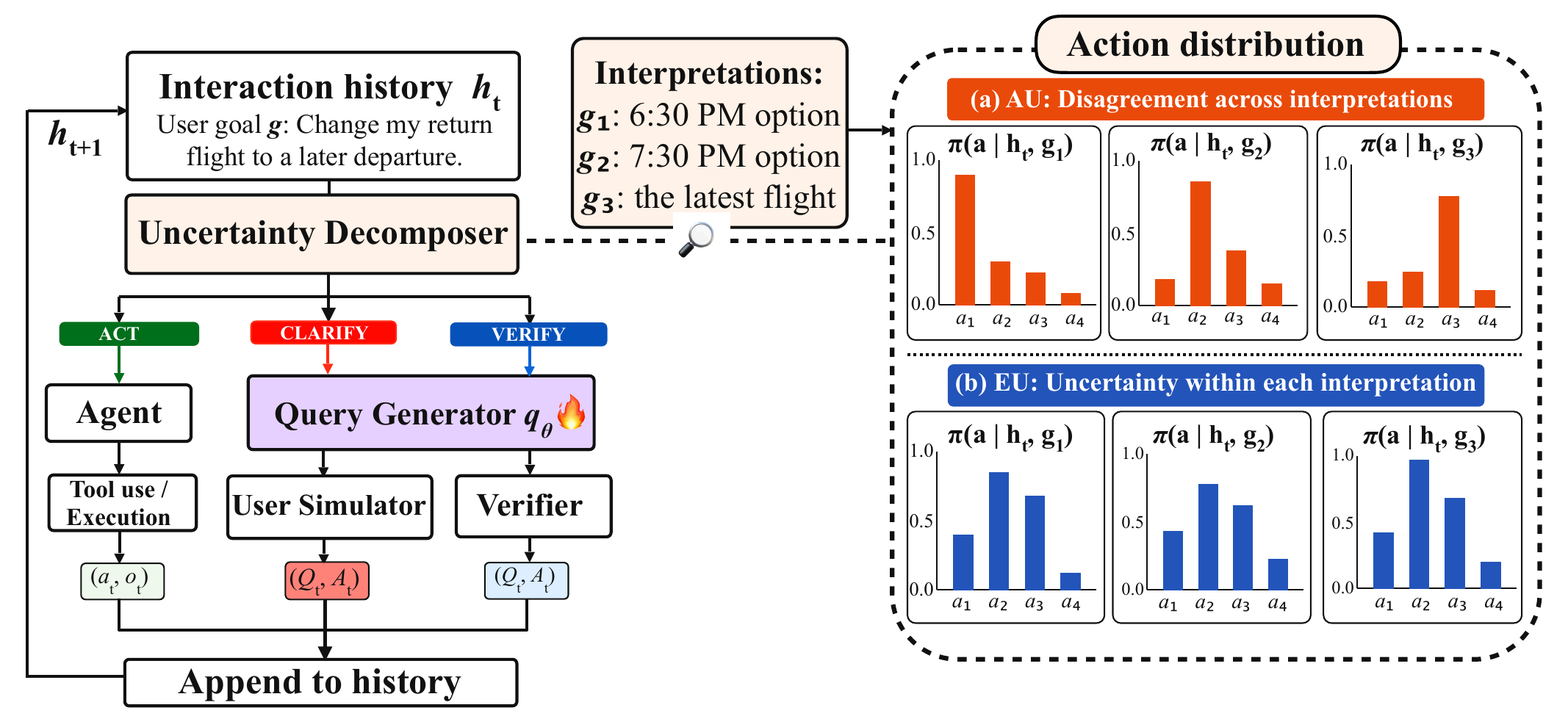}
        \caption{Uncertainty Decomposition and Routing}
        \label{fig:pipeline}
    \end{subfigure}%
    \begin{subfigure}[t]{0.33\textwidth}
    \setlength{\abovecaptionskip}{3pt}
    \setlength{\belowcaptionskip}{2pt}
        \centering
        \includegraphics[width=\linewidth]{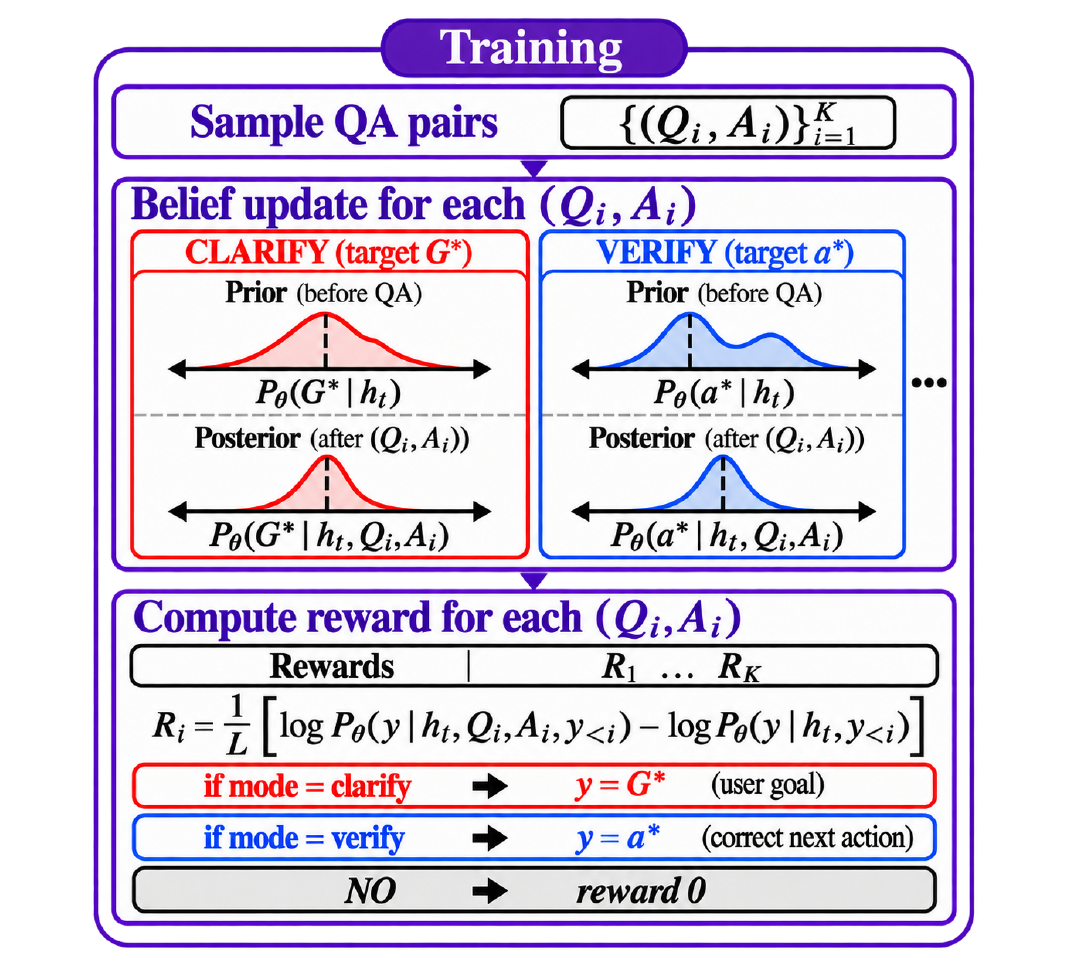}
        \caption{Mode-Conditioned Information-Gain Training}
        \label{fig:dapo}
    \end{subfigure}

    \caption{\textbf{Overview of the \textsc{Prour} framework}.
    (a) Decomposer estimates action-level aleatoric and epistemic uncertainty
    from candidate user-goal interpretations and routes to
    \textsc{clarify}, \textsc{verify}, or \textsc{act}.
    (b) Query Generator is trained with a mode-conditioned
    information-gain reward, using the ground-truth user goal as the clarification target and reference next action as the verification target.}
    \label{fig:pipeline_dapo}
    \vspace{-4mm}
\end{figure}

When an agent is uncertain about its next action, it must answer two questions: \emph{where does the missing information reside}, and \emph{what information should be acquired from that source}? As illustrated in Figure~\ref{fig:pipeline_dapo}, the \textsc{Prour} framework addresses these questions with two components. First, a \textbf{Decomposer} separates uncertainty induced by ambiguity in the user's goal from uncertainty that remains after ambiguity is clarified, and accordingly routes the agent among \textsc{clarify}, \textsc{verify}, and \textsc{act}. Second, when intervention is required, a shared \textbf{Query Generator} acquires information from the routed source: the user under \textsc{clarify}, or grounded world evidence under \textsc{verify}.

\subsection{Problem Formulation}
\label{sec:source-aware-interaction}

We study uncertainty resolution in 
tool-using agents \citep{yao2022react,schick2023toolformer,yao2024taubench}, where an agent interacts with a user and invokes domain tools. 
At each decision step, the agent may already have sufficient information to act, or its next action may depend on information unavailable in the current history. Importantly, such information can reside in two different sources: \emph{user-side} information, such as an underspecified preference or intent, and \emph{world-side} information, such as environment state, entity attributes, policy constraints, or action validity. The agent must therefore choose among three modes:
\textsc{act} using the available information,
\textsc{clarify} by acquiring information from the user, or
\textsc{verify} by acquiring grounded evidence about the world. We represent an interaction trajectory $\tau$ and the history available before decision step $t$ as
\begin{equation}
    \tau
    = \left(U,\left(a_i,o_i,Q_i,A_i\right)_{i=1}^{T},G\right),
    \qquad
    h_t
    = \left(U,\left(a_i,o_i,Q_i,A_i\right)_{i=1}^{t-1}\right).
    \label{eq:history}
\end{equation}
Here, $U$ is the initial user message and $G$ is the latent user goal hidden from the agent during interaction. $(Q_i,A_i)$ is a query--response pair, and $(a_i,o_i)$ is a tool interaction. At step $i$, the decomposer selects a mode $m_i\in\{\texttt{act},\texttt{clarify},\texttt{verify}\}$. Under \texttt{act} mode, the agent selects
$a_i\in\mathcal{A}=\mathcal{A}_{\mathrm{tool}}\cup\{\texttt{respond} \}$ where a tool action produces observation $o_i$ with $Q_i=A_i=\varnothing$. Under either intervention mode, the query generator issues a query $Q_i$ and receives a response $A_i$ by either clarifying the user or verifying the world, with $a_i=o_i=\varnothing$. \textsc{act} may itself include ordinary tool calls when the appropriate next action is sufficiently determined from the current history, whereas \textsc{verify} is used when missing world-side evidence leaves the appropriate next action uncertain. Policy-mandated user confirmation is a \texttt{respond} action issued under \texttt{act} mode.

\subsection{Uncertainty Decomposition and Routing}
\label{sec:goal-conditioned-decomposition}

As illustrated in Figure~\ref{fig:pipeline}, at each interaction step, the decomposer determines whether uncertainty about the next action originates from the user's latent goal or from insufficient knowledge of the world. We adapt the distinction between aleatoric and epistemic uncertainty to the agent's \emph{action space}. Specifically, \textbf{Aleatoric Uncertainty (AU)} captures disagreement in the
predicted next-action distributions across plausible interpretations
of the user's goal; high AU therefore indicates user-side ambiguity that should be resolved through \textsc{clarify}. In contrast, \textbf{Epistemic Uncertainty (EU)} captures the residual uncertainty over the next action that remains under a fixed goal interpretation; high EU indicates missing world-side evidence---such as environment state, entity grounding, policy constraints, calculation results, or execution status---that should be resolved through \textsc{verify}.

\textbf{Goal-Conditioned Action Distribution.} At decision $t$, the decomposer is invoked $K$ times, and
each call independently proposes one interpretation $g_k$ of the user's
latent goal conditioned on the observed history $h_t$, without access to the other $K-1$ samples. Notably, these candidate goal interpretations are hypotheses inferred from the observed history \(h_t\), instead of perturbations or samples of the ground-truth goal $G$, which is not provided to the decomposer or agent. As detailed in Appendix~\ref{app:routing-calibration}, the decomposer then samples structured
candidate actions $a$ containing a tool name and its arguments, and evaluates action distribution under each interpretation $g_k$: 
\begin{equation}
    p_{t,k}(a) = p_{\phi}(a\mid h_t,g_k),\quad
    \bar p_t(a) = \frac{1}{K}\sum\nolimits_{k=1}^{K}p_{t,k}(a).
\label{eq:goal-conditioned-action-distribution}
\end{equation}
where $\bar p_t$ is the empirical mixture across sampled interpretations.

\textbf{Information-Theoretic Decomposition.} Let $\widetilde G_t$ denote the empirical random variable assigning equal probability to each sampled candidate interpretations $\{g_k\}_{k=1}^{K}$, and let $a_t\in\mathcal{A}$ denote the predicted
next-action variable. By conditional mutual information~\citep{cover2006elements}:
\begin{equation}
    H(a_t\mid h_t)
    =
    I(a_t;\widetilde G_t\mid h_t)
    +
    H(a_t\mid \widetilde G_t,h_t).
    \label{eq:conditional-mi-decomposition}
\end{equation}
where $I(\cdot;\cdot)$ denotes mutual information, and $H(p)=-\sum_{a\in\mathcal{A}}p(a)\log p(a)$ denotes Shannon entropy.
Based on the empirical sampled mixture in Eq.~\ref{eq:goal-conditioned-action-distribution}, we estimate aleatoric uncertainty (AU) and epistemic uncertainty (EU) as follows:
\begin{align}
    \mathrm{AU}_t
    &:= I(a_t;\widetilde G_t\mid h_t)
    =
    H(\bar p_t)
    -
    \frac{1}{K}\sum\nolimits_{k=1}^{K} H(p_{t,k})
    =
    \frac{1}{K}\sum\nolimits_{k=1}^{K}
    D_{\mathrm{KL}}\!\left(p_{t,k}\,\|\,\bar p_t\right),
    \label{eq:au-score}\\
    \mathrm{EU}_t
    &:= \frac{1}{K}\sum\nolimits_{k=1}^{K}H(p_{t,k}).
      \label{eq:eu-score}
\end{align}

We use the empirical mutual-information in Eq.~\ref{eq:au-score} as AU, because it measures disagreement across plausible candidate interpretations. Mutual information equals the average KL divergence between each interpretation-conditioned action distribution and their empirical mixture, so it is zero when all $p_{t,k}$ are identical, and increases as the action distributions diverge from their average $\bar p_t$. Thus, under our formulation, high $\mathrm{AU}$ means that different plausible interpretations of the latent goal imply different next actions, indicating user-side ambiguity that should be resolved by clarification. 

Similarly, we use the empirical conditional-entropy in Eq.~\ref{eq:eu-score} as EU. This is the average entropy of the action distribution after conditioning on an interpretation. It is therefore high when the agent remains uncertain over the next action even after user-side ambiguity has been removed, indicating missing world-side evidence that should be resolved by verification. Consequently,  $H(\bar p_t)=\mathrm{AU}_t+\mathrm{EU}_t$.

\textbf{Routing Policy.}
We convert the decomposition into a routing decision, as illustrated in Figure~\ref{fig:pipeline}:
\begin{equation}
    m_t =
    \begin{cases}
        \texttt{clarify}, & \text{if} \ \mathrm{AU}_t>\tau_{\mathrm{au}},\\
        \texttt{verify},  & \text{else if} \ \mathrm{AU}_t\leq\tau_{\mathrm{au}}
                             \ \text{and}\ 
                             \mathrm{EU}_t>\tau_{\mathrm{eu}},\\
        \texttt{act},     & \text{otherwise}.
    \end{cases}
    \label{eq:routing}
\end{equation}
The ordering is intentional. When substantial user-side ambiguity remains, the relevant world-side query may itself depend on which goal interpretation is correct. The system therefore resolves user-side uncertainty first, while verification is triggered only when the goal is sufficiently determined but uncertainty about the next action remains. When neither uncertainty exceeds its calibrated threshold, the agent proceeds directly without unnecessary intervention. $\tau_{\mathrm{au}}$ and $\tau_{\mathrm{eu}}$ are tuned on a development set disjoint from training and evaluation.

\subsection{Mode-conditioned Information-Gain Reward}
\label{sec:mode-conditioned-ig}

The decomposer determines \emph{where} missing information should be acquired, but not whether the generated query is effective at resolving the underlying uncertainty. Building on the formulation of information gain in~\citet{deng2026uncertainty}, we introduce an information-gain reward to train a shared query generator for both clarification and verification, rather than learning separate policies for each mode. Figure~\ref{fig:dapo} summarizes the overall training procedure. The routing mode determines both the information source and the target used to compute the reward: clarification queries the user simulator to reduce uncertainty about the ground-truth user goal, whereas verification queries a verifier to reduce uncertainty about the reference next action.

\textbf{Mode-Conditioned Information Acquisition.} The query generator samples
\begin{equation}
    Q_t\sim q_{\theta}(\cdot\mid h_t,m_t).
    \label{eq:mode-conditioned-query-policy}
\end{equation}
where mode $m_t\in\{\texttt{clarify},\texttt{verify}\}$.
Clarification queries aim at making the ground-truth user goal more identifiable, while verification queries aim at eliciting evidence to identify the reference next action. Therefore, response source $A_t$ and reward target $Y_t$ are conditioned on the mode:
\begin{equation}
(A_t,Y_t)=
\begin{cases}
\bigl(\mathcal{S}_{\mathrm{user}}(Q_t\mid h_t,G^*),\;G^*\bigr),
& m_t=\texttt{clarify},\\[2pt]
\bigl(\mathcal{V}(Q_t\mid h_t,\mathcal{W},\mathcal{T}_{\mathrm{tool}}),\;a_t^*\bigr),
& m_t=\texttt{verify}.
\end{cases}
\label{eq:mode-conditioned-source-target}
\end{equation}
Here, $G^*$ is the normalized ground-truth user goal and $a_t^*$ is the next
reference tool call in the $\tau$-bench environment. This formulation does not assume that \(a_t^*\) is a universal or uniquely valid action; rather, it provides a consistent supervision target available from the reference trajectory. Responses $A_t$ come from the user simulator $\mathcal{S}_{\mathrm{user}}$ and verifier $\mathcal{V}$.  The verifier may retrieve environment state, check domain policy $\mathcal{W}$ and arguments of read-only tools $\mathcal{T}_{\mathrm{tool}}$,  or assess whether the requested subgoals have been completed. Since the verifier
cannot use tools that modify the environment, verification does not change the task state. The completed query--response pair is added to history. 

\textbf{Information-Gain Reward.}
We define the mode-conditioned objective so an optimal query maximizes the \emph{Expected Information Gain (EIG)}~\citep{deng2026uncertainty}:
\begin{equation}
\begin{aligned}
    Q_t^*
    &= \arg\max\nolimits_Q I(Y_t;A_t\mid Q,h_t,m_t)\\
    &= \arg\max\nolimits_Q
      \mathbb{E}_{A_t\sim P_{m_t}(\cdot\mid Q,h_t)}
      \left[H(Y_t\mid h_t,m_t)
      -H(Y_t\mid h_t,m_t,Q,A_t)\right],
\end{aligned}
\label{eq:mode-conditioned-eig}
\end{equation}
where $P_{m_t}$ is the response distribution for mode $m_t$.
Exact EIG is intractable for open-ended queries and responses, so we adopt the teacher-forced belief-update approximation of \citet{deng2026uncertainty} and apply it to the
mode-conditioned target. Let $\ell_{\theta}(Y\mid c)$ denote the
length-normalized teacher-forced log-likelihood of target $Y$ under context $c$.
The information-gain reward is the shift from prior belief to posterior belief after incorporating the query--response pair:
\begin{equation}
    R_t
    = \ell_{\theta}(Y_t\mid h_t,m_t,Q_t,A_t)
      -\ell_{\theta}(Y_t\mid h_t,m_t).
    \label{eq:mode-conditioned-ig-reward}
\end{equation}
A positive reward indicates that the query-response pair $(Q_t,A_t)$ increases the likelihood of the correct target, whereas a negative reward indicates that it makes the target less identifiable. If no query is issued, we
set $R_t=0$.

\textbf{On-Policy Training.} For each state-mode pair $(h_t,m_t)$, the rollout
policy samples a group of candidate queries. Clarification candidates receive
responses from a strict user simulator conditioned on $G^*$, whereas verification
candidates receive responses from the verifier; both responder LLMs are frozen.
The strict user simulator reveals task-specific information only for precise,
relevant queries, while the verifier only returns evidence supported by its
grounded context. Each query--response pair receives the reward in
Eq.~\ref{eq:mode-conditioned-ig-reward}, and the query generator is
optimized via Decoupled Clip and Dynamic Sampling Policy Optimization (DAPO) \citep{yu2025dapo}. All responder LLMs, the agent, and decomposer remain frozen; only the query generator is optimized.
The belief update, rollout estimator, advantage computation, and DAPO
objective are given in Appendix~\ref{app:dapo-details}.

\begin{figure}[t]
\vspace{-3mm}
    \centering
    \begin{subfigure}[t]{0.495\textwidth}
        \centering
        \includegraphics[width=\linewidth]{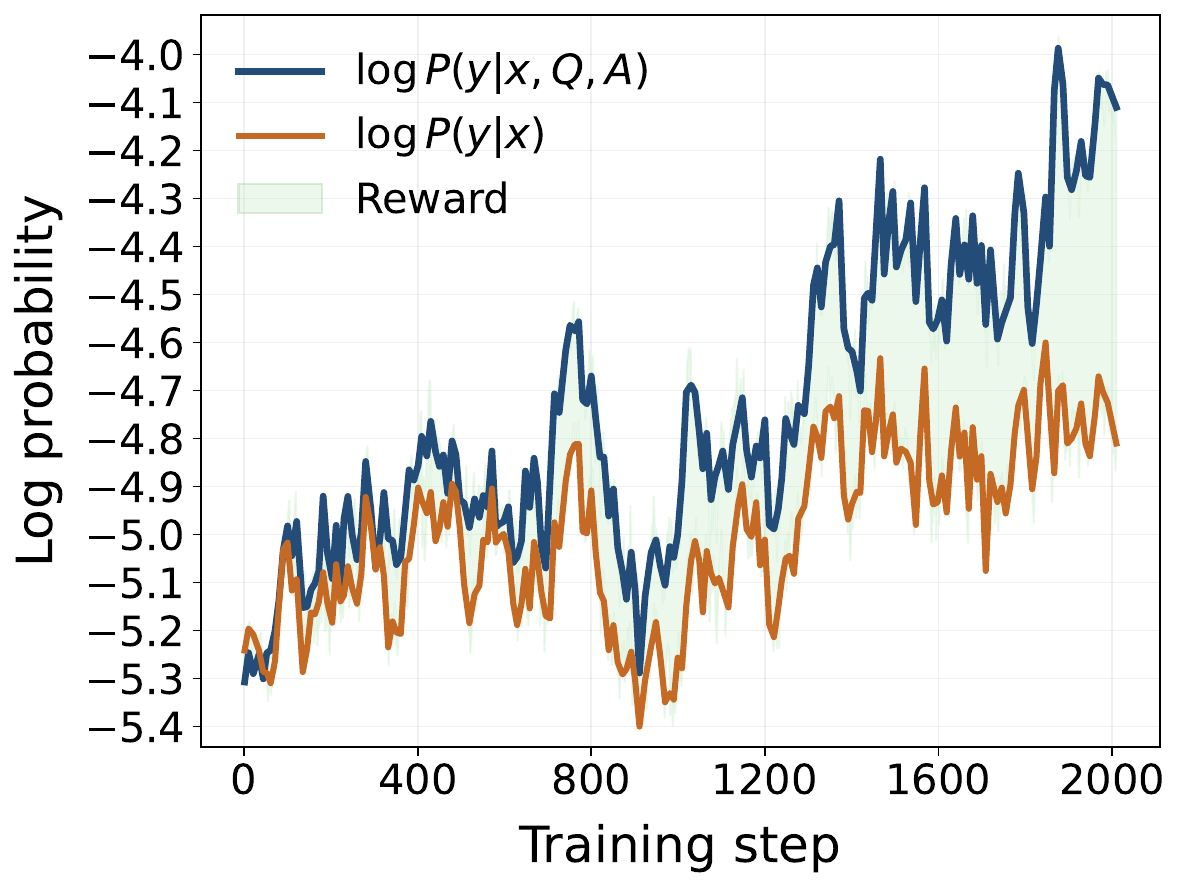}
        \caption{Prior and posterior log-likelihoods of the
        mode-conditioned target $y$.}
        \label{fig:log-prob}
    \end{subfigure}
    \begin{subfigure}[t]{0.495\textwidth}
        \centering
        \includegraphics[width=\linewidth]{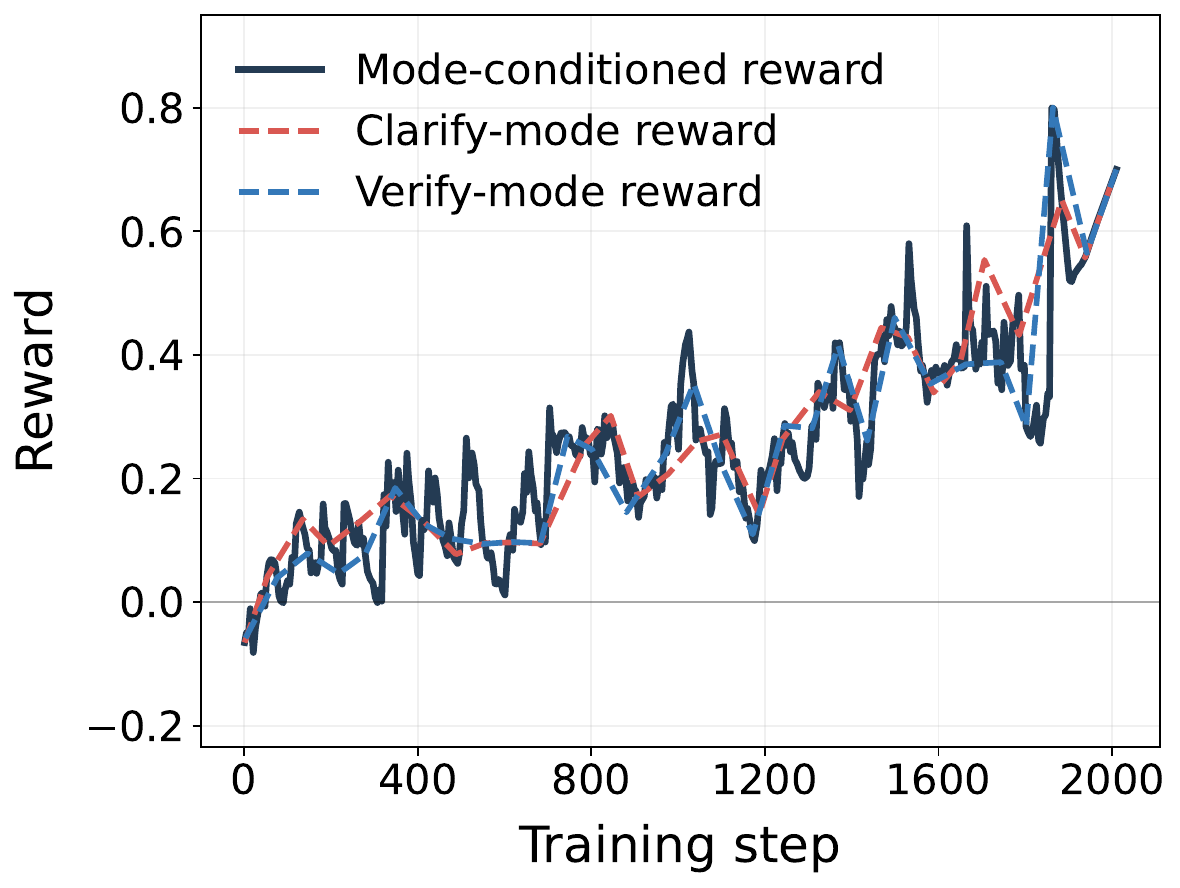}
        \caption{Mean information-gain reward over all samples and separate rewards over \textsc{clarify}- and \textsc{verify}-mode.}
        \label{fig:reward}
    \end{subfigure}
    \setlength{\abovecaptionskip}{2pt}
    \caption{\textbf{Training dynamics under mode-conditioned information-gain reward.}}
    \label{fig:training-dynamics}
    \vspace{-5mm}
\end{figure}

\section{Experiments}
\label{sec:experiments}

We evaluate \textsc{Prour} in the $\tau$-bench environment \citep{yao2024taubench}. Following previous studies \citep{deng2026uncertainty}, we train exclusively on 500
$\tau$-retail trajectories and evaluate on 115 held-out retail tasks and 50
airline tasks. Airline serves as the out-of-distribution domain and is excluded
from both training and calibration. 
We additionally evaluate cross-benchmark generalization on the transactional text domains of $\tau^3$-bench \citep{barres2025tau2}---retail, airline, and telecom, without retraining.
Qwen3-1.7B serves as the query generator and is the only optimized module. By default, Qwen3-8B implements all frozen
modules: the decomposer, agent, user simulator, and verifier.
We compare against IG-clarifier~\citep{deng2026uncertainty},
Ask-before-Plan~\citep{zhang2024askbeforeplan}, VerifiAgent~\citep{han2025verifiagent}, SABER~\citep{cuadron2025saber}, SAGE-Agent~\citep{suri2026structured}, and controlled ablations, reporting pass rates and average steps. Appendix~\ref{app:experimental-details}
provides the implementation details and hyperparameters. Appendix~\ref{app:prompt-templates} provides the prompt templates.

\subsection{Training Dynamics and Routing Calibration}
\label{sec:training-routing}

\paragraph{Training Dynamics.} To characterize how the query generator evolves under the mode-conditioned reward, we track the likelihood and reward throughout training.
Figure~\ref{fig:training-dynamics}(a) compares the prior belief
$\ell_{\theta}(Y_t\mid h_t,m_t)$ with the posterior belief $\ell_{\theta}(Y_t\mid h_t,m_t,Q_t,A_t)$. Both curves
exhibit mild oscillations because of variation in sampled
contexts and mode-conditioned targets. Nevertheless, $\ell_{\mathrm{post}}$ becomes consistently
higher than $\ell_{\mathrm{prior}}$, and their gap generally grows over
training. This suggests that the learned policy provides useful evidence for identifying the mode-conditioned target, rather than merely raising the model's prior confidence before a query is generated.

Consistent with this interpretation, Figure~\ref{fig:training-dynamics}(b) reports the corresponding reward
$R_t=\ell_{\mathrm{post}}-\ell_{\mathrm{prior}}$ by showing both the aggregate reward across all samples and separated into \textsc{clarify} and \textsc{verify} mode. The mean reward rises from near zero
to approximately $0.7$ after 2,010 optimization steps, and the \textsc{clarify}- and
\textsc{verify}-mode rewards improve alongside the aggregate reward without evidence of mode collapse. These dynamics
show that a shared query generator can optimize two mode-conditioned reward targets while maintaining a stable learning signal.
\paragraph{Routing Calibration.}
Because AU and EU operate on different empirical scales, we calibrate their
routing thresholds on a disjoint $\tau$-retail development set of 240 decision steps, where each step is labeled as \textsc{AU}, \textsc{EU}, or
\textsc{No\_uncertainty}. The decomposer then samples $5$ goal interpretations for each step to estimate AU and EU, and a grid-search procedure selects the thresholds that maximize three-class macro-F1, yielding $(\tau_{\mathrm{AU}},\tau_{\mathrm{EU}})=(0.25,0.20)$. We retain these thresholds for all main experiments, including Section~\ref{sec:cross-agent}. Appendix~\ref{app:routing-calibration} details the calibration protocol, threshold search, the confusion matrix, and a separate recalibration study on GLM-4-32B.

{
\setlength{\textfloatsep}{1pt}
\setlength{\abovecaptionskip}{1pt}
\setlength{\belowcaptionskip}{1pt}
\begin{table}[t]
\centering
\small
\caption{
Main results on $\tau$-bench. We report Pass@1 (\%) and average interaction steps. \textit{Avg. Clarify} and \textit{Avg. Verify} count routed \textsc{clarify} and \textsc{verify} steps issued as separate turns; methods that interleave them with agent turns are counted under \textit{Avg. Act}. \textit{LLM Router}, \textit{Clarify-only}, and \textit{Verify-only} are controlled ablations.
}
\label {tab:main_results}
\setlength{\tabcolsep}{3.5pt}
\renewcommand{\arraystretch}{1.12}
\begin{adjustbox}{max width=\textwidth}
\begin{tabular}{lccccccc}
\toprule
\multicolumn{1}{c}{\textbf{Method}} &
\multicolumn{3}{c}{\textbf{Success Rate (\% $\uparrow$)}} &
\multicolumn{4}{c}{\textbf{Steps $\downarrow$}} \\
\cmidrule(lr){1-1}
\cmidrule(lr){2-4}
\cmidrule(lr){5-8}

\textbf{Strategy} &
\textbf{Retail} &
\textbf{Airline} &
\textbf{Average} &
\textbf{Avg. \textsc{act}} &
\textbf{Avg. \textsc{clarify}} &
\textbf{Avg. \textsc{verify}} &
\textbf{Total} \\
\midrule

Base Agent
& 16.80
& 18.00
& 17.40
& 28.90
& --
& --
& 28.90 \\

Ask-before-Plan \citep{zhang2024askbeforeplan}
& 23.19
& 24.00
& 23.60
& 21.10
& 2.40
& --
& 23.50 \\

VerifiAgent \citep{han2025verifiagent}
& 19.13
& 26.00
& 22.57
& 31.81
& --
& 11.20
& 43.01 \\

SABER \citep{cuadron2025saber}
& 21.74 & 20.00 & 20.87
& 34.43 & -- & -- & 34.43 \\

SAGE-Agent \citep{suri2026structured}
& 18.26 & 24.00 & 21.13
& 32.65 & 1.89 & -- & 34.54 \\

IG-clarifier \citep{deng2026uncertainty}
& 21.70
& 24.00
& 22.85
& 20.85
& 2.70
& --
& 23.55 \\

\midrule

LLM Router
& 23.50
& 28.00
& 25.75
& 16.59
& 3.30
& 3.50
& 23.39 \\

Clarify-only
& 24.30
& 26.00
& 25.15
& 21.17
& 2.03
& --
& 23.20 \\

Verify-only
& 15.70
& \textbf{30.00}
& 22.85
& 22.22
& --
& 8.48
& 30.70 \\

\rowcolor{blue!10}
\textbf{\textsc{Prour}}
& \textbf{27.00}
& 29.33
& \textbf{28.17}
& 14.34
& 2.20
& 4.79
& \textbf{21.33} \\

\bottomrule

\end{tabular}
\end{adjustbox}
\end{table}
}

{
\setlength{\textfloatsep}{1pt}
\setlength{\abovecaptionskip}{1pt}
\setlength{\belowcaptionskip}{1pt}
\begin{table}[t]
\centering
\small
\caption{
Effect of the query generator on $\tau$-bench. All alternative generators are prompt-only; the decomposer, task agent, and routing thresholds are held fixed.
}
\label {tab:query-generators}
\setlength{\tabcolsep}{3.5pt}
\renewcommand{\arraystretch}{1.12}
\begin{adjustbox}{max width=\textwidth}
\begin{tabular}{lccccccc}
\toprule
\multicolumn{1}{c}{\textbf{Method}} &
\multicolumn{3}{c}{\textbf{Success Rate (\% $\uparrow$)}} &
\multicolumn{4}{c}{\textbf{Steps $\downarrow$}} \\
\cmidrule(lr){1-1}
\cmidrule(lr){2-4}
\cmidrule(lr){5-8}

\textbf{Query Generator} &
\textbf{Retail} &
\textbf{Airline} &
\textbf{Average} &
\textbf{Avg. \textsc{act}} &
\textbf{Avg. \textsc{clarify}} &
\textbf{Avg. \textsc{verify}} &
\textbf{Total} \\
\midrule

None (Base Agent)
& 16.80 & 18.00 & 17.40 & 28.90 & -- & -- & 28.90 \\

Qwen3-1.7B
& 24.35 & 20.67 & 22.51 & 18.51 & 3.86 & 5.11 & 27.48 \\

Qwen3-8B
& 26.09 & 22.00 & 24.05 & 12.45 & 2.37 & 5.62 & \textbf{20.44} \\

Qwen3-32B
& 28.70 & 24.00 & 26.35 & 17.75 & 1.58 & 7.22 & 26.55 \\

DeepSeek-V4-Flash
& \textbf{29.60} & 28.00 & \textbf{28.80} & 13.17 & 2.42 & 5.94 & 21.53 \\

\rowcolor{blue!10}
\textbf{\textsc{Prour} (Qwen3-1.7B-trained)}
& 27.00 & \textbf{29.33} & 28.17 & 14.34 & 2.20 & 4.79 & 21.33 \\

\bottomrule
\end{tabular}
\end{adjustbox}
\vspace{-3mm}
\end{table}
}

\subsection{Main Results and Ablation Studies}
\label{sec:main-results}

\textbf{End-to-end Performance.}~
Table~\ref{tab:main_results} shows that \textsc{Prour} achieves the highest average
success rate ($28.17\%$) while requiring the fewest total interaction steps
(21.33), compared with $17.40\%$ success and 28.90 steps for the base agent.
Ask-before-Plan, SAGE-Agent, and IG-clarifier seek information from the user but do not acquire
grounded world-side evidence, attaining average success rates of 
$23.60\%$, $21.13\%$, and $22.85\%$, respectively. VerifiAgent instead performs meta- and tool-based
verification without explicitly routing user-side uncertainty; despite its
strong airline result, it requires 43.01 steps on average. SABER requests user consent on mutating actions and achieves $20.87\%$ success on average. These results
show that resolving both user-side and world-side uncertainty through targeted routing is
more effective than relying exclusively on clarification or verification.

To isolate the contribution of routing and information source, the LLM Router
replaces uncertainty decomposition with a prompt-only decision, while the
Clarify-only and Verify-only variants disable one intervention branch. Each
variant underperforms the full method on average. In particular, Verify-only
reaches $30.0\%$ on airline but only $15.7\%$ on retail, whereas Clarify-only
is more balanced across domains. This contrasting behavior demonstrates the benefit of routing
between user-side and world-side evidence rather than applying either
intervention uniformly. 

\textbf{Effect of the Query Generator.}~
\label{sec:query-generator}
To isolate the contribution of the learned query-generation policy, we
hold the task agent, decomposer, routing thresholds, and all other auxiliary
modules fixed, and vary only the query generator. As shown in
Table~\ref{tab:query-generators}, the trained Qwen3-1.7B generator achieves
$28.17\%$ average success, outperforming the prompted Qwen3-1.7B, Qwen3-8B, and Qwen3-32B
generators and approaching DeepSeek-V4-Flash ($28.80\%$) with slightly fewer steps. This suggests that the improvement does not simply arise from increasing module capacity. Instead, training specializes the generator toward queries that acquire information useful for resolving uncertainty, demonstrating the benefit of explicitly optimizing query generation for uncertainty resolution.

{
\setlength{\textfloatsep}{1pt}
\setlength{\abovecaptionskip}{1pt}
\setlength{\belowcaptionskip}{1pt}
\begin{table}[t]
\centering
\small
\caption{
Cross-agent generalization performance on $\tau$-bench.
Different agent backbones are compared under varying query generators. The best results are bolded.
}
\label{tab:cross-agent}
\setlength{\tabcolsep}{3.5pt}
\renewcommand{\arraystretch}{1.12}
\begin{adjustbox}{max width=\textwidth}
\begin{tabular}{llccccccc}
\toprule
\multicolumn{2}{c}{\textbf{Setting}}
& \multicolumn{3}{c}{\textbf{Success Rate (\% $\uparrow$)}}
& \multicolumn{4}{c}{\textbf{Steps $\downarrow$}} \\
\cmidrule(r){1-2}
\cmidrule(lr){3-5}
\cmidrule(l){6-9}

\textbf{Agent Model}
& \textbf{Query Generator}
& \textbf{Retail}
& \textbf{Airline}
& \textbf{Average}
& \textbf{Avg. \textsc{act}}
& \textbf{Avg. \textsc{clarify}}
& \textbf{Avg. \textsc{verify}}
& \textbf{Total} \\
\midrule

\multirow{3}{*}{Qwen3-8B}
& None
& 16.80
& 18.00
& 17.40
& 28.90
& --
& --
& 28.90 \\

& Qwen3-8B
& 26.09
& 22.00
& 24.05
& 12.45
& 2.37
& 5.62
& \textbf{20.44} \\

\rowcolor{blue!10}
& \textbf{\textsc{Prour}}
& \textbf{27.00}
& \textbf{29.33}
& \textbf{28.17}
& 14.34
& 2.20
& 4.79
& 21.33 \\
\midrule

\multirow{3}{*}{Qwen3-32B}
& None
& 26.38
& 24.00
& 25.19
& 22.42
& --
& --
& 22.42 \\

& Qwen3-8B
& 31.30
& \textbf{46.00}
& 38.65
& 12.19
& 1.57
& 6.08
& 19.84 \\

\rowcolor{blue!10}
& \textbf{\textsc{Prour}}
& \textbf{36.50}
& 44.00
& \textbf{40.25}
& 11.23
& 1.53
& 6.11
& \textbf{18.87} \\
\midrule

\multirow{3}{*}{DeepSeek-V4-Flash}
& None
& 36.23
& 38.00
& 37.12
& 32.30
& --
& --
& 32.30 \\

& Qwen3-8B
& 47.80
& \textbf{46.00}
& 46.90
& 15.17
& 1.65
& 7.85
& \textbf{24.67} \\

\rowcolor{blue!10}
& \textbf{\textsc{Prour}}
& \textbf{50.40}
& \textbf{46.00}
& \textbf{48.20}
& 14.48
& 2.57
& 7.79
& 24.84 \\

\bottomrule
\end{tabular}
\end{adjustbox}
\end{table}
}

{
\setlength{\textfloatsep}{1pt}
\setlength{\abovecaptionskip}{1pt}
\setlength{\belowcaptionskip}{1pt}
\begin{table}[t]
\centering
\small
\caption{
Cross-benchmark generalization to $\tau^3$-bench.
All modules and routing thresholds are the same as $\tau$-bench. \textit{Avg. Clarify} and \textit{Avg. Verify} count routed \textsc{clarify} and \textsc{verify} steps issued as separate turns; methods that interleave them with agent turns are counted under \textit{Avg. Act}.
}
\label{tab:tau3-results}
\setlength{\tabcolsep}{3.5pt}
\renewcommand{\arraystretch}{1.12}
\begin{adjustbox}{max width=\textwidth}
\begin{tabular}{lcccccccc}
\toprule
\multicolumn{1}{c}{\textbf{Method}} &
\multicolumn{4}{c}{\textbf{Success Rate (\% $\uparrow$)}} &
\multicolumn{4}{c}{\textbf{Steps $\downarrow$}} \\
\cmidrule(lr){1-1}
\cmidrule(lr){2-5}
\cmidrule(lr){6-9}

\textbf{Strategy} &
\textbf{Retail} &
\textbf{Airline} &
\textbf{Telecom} &
\textbf{Average} &
\textbf{Avg. \textsc{act}} &
\textbf{Avg. \textsc{clarify}} &
\textbf{Avg. \textsc{verify}} &
\textbf{Total} \\
\midrule

Base Agent
& 17.25 & 20.00 & 13.70 & 16.98
& 33.68 & -- & -- & 33.68 \\

Ask-before-Plan \citep{zhang2024askbeforeplan}
& 24.60 & 32.00 & 21.90 & 26.17
& 33.57 & 2.95 & -- & 36.52 \\

VerifiAgent \citep{han2025verifiagent}
& 26.30 & 32.00 & 23.39 & 27.23
& 19.05 & -- & 11.79 & 30.84 \\

SABER \citep{cuadron2025saber}
& 21.64 & 27.33 & 21.90 & 23.62
& 44.03 & -- & -- & 44.03 \\

SAGE-Agent \citep{suri2026structured}
& 19.30 & 28.00 & 24.56 & 23.95
& 35.51 & 1.78 & -- & 37.29 \\

IG-clarifier \citep{deng2026uncertainty}
& 24.00 & 26.00 & 22.81 & 24.27
& 27.95 & 2.80 & -- & 30.75 \\

\midrule

LLM Router
& 21.90 & \textbf{36.00} & 21.10 & 26.33
& 24.29 & 1.33 & 4.72 & 30.34 \\

Clarify-only
& 26.30 & 28.00 & 20.20 & 24.83
& 29.54 & 2.75 & -- & 32.29 \\

Verify-only
& 25.40 & 28.00 & 25.40 & 26.27
& 20.90 & -- & 12.57 & 33.47 \\

\rowcolor{blue!10}
\textbf{\textsc{Prour}}
& \textbf{29.80}
& 30.00
& \textbf{27.20}
& \textbf{29.00}
& 18.46
& 1.60
& 9.19
& \textbf{29.25} \\

\bottomrule
\end{tabular}
\end{adjustbox}
\vspace{-3mm}
\end{table}
}

\begin{figure}[t]
    \centering
    \begin{subfigure}[t]{0.495\textwidth}
        \centering
        \includegraphics[width=\linewidth]{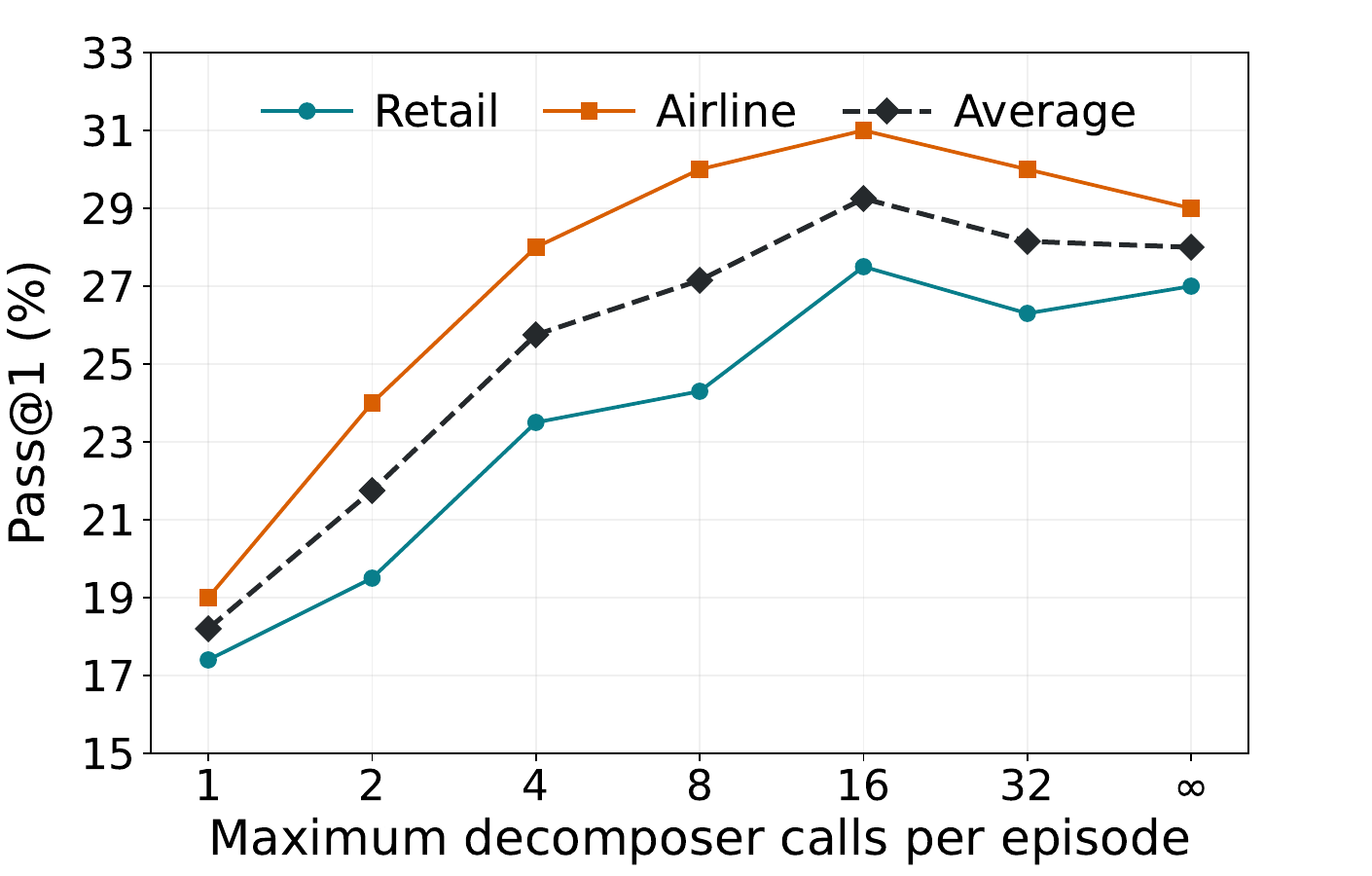}
        \caption{Task success w.r.t. decomposer-call budgets.}
        \label{fig:decomposer-budget-success}
    \end{subfigure}
    \begin{subfigure}[t]{0.495\textwidth}
        \centering
        \includegraphics[width=\linewidth]{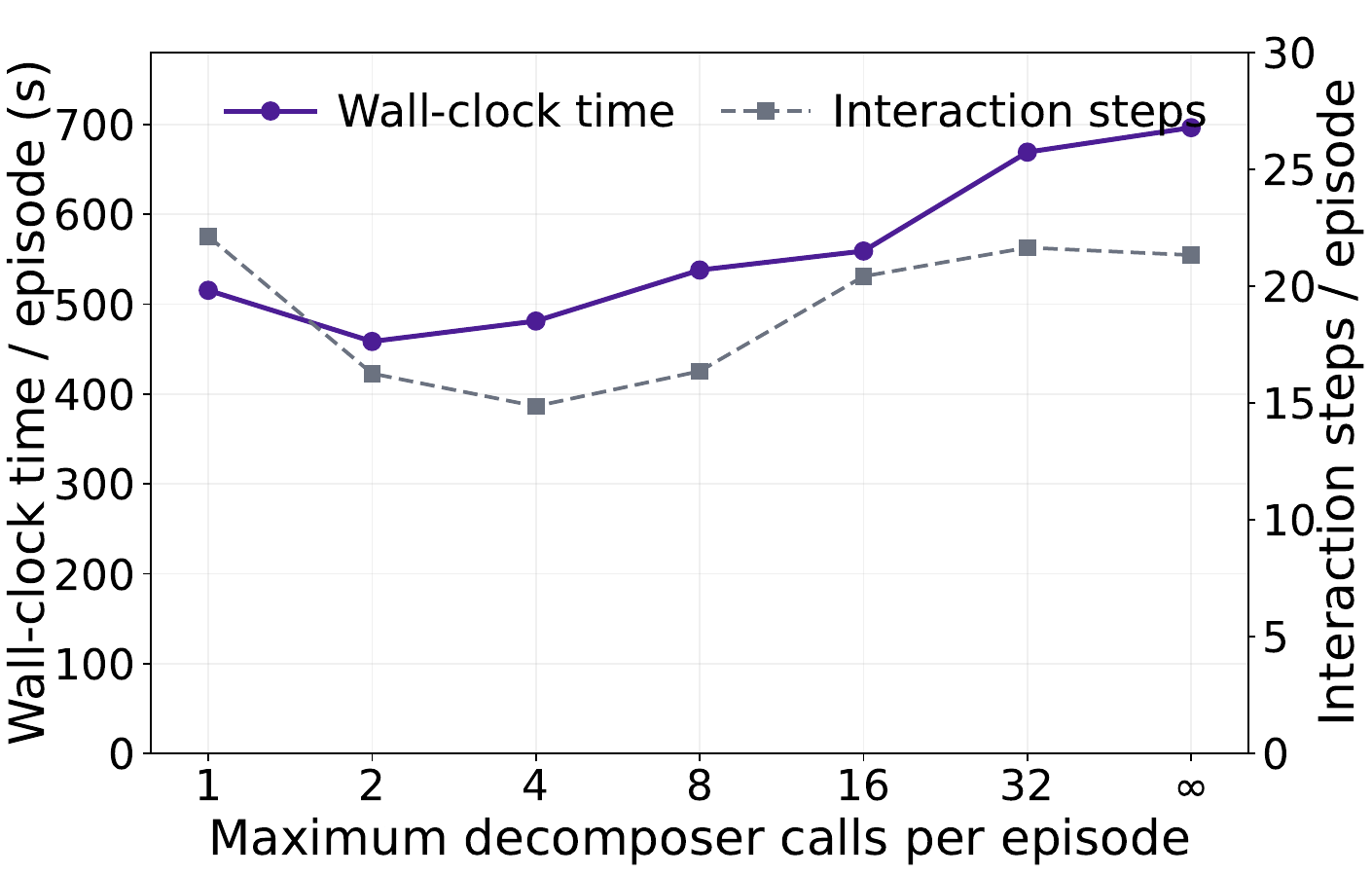}
        \caption{Average wall-clock time and interaction steps.}
        \label{fig:decomposer-budget-cost}
    \end{subfigure}
    \setlength{\abovecaptionskip}{4pt}
    \caption{\textbf{Effect of the decomposer-call budget on performance and
    cost.} (a) Success rate on retail, airline, and their average.
    (b) Average wall-clock time and total interaction steps per episode. The
    $\infty$ setting imposes no per-episode limit on decomposer calls.}
    \label{fig:decomposer-budget}
    \vspace{-6mm}
\end{figure}

\subsection{Generalization Across Agents and Benchmarks}
\label{sec:generalization}

\textbf{Cross-agent Generalization.}~
\label{sec:cross-agent}
To assess whether the learned policy generalizes across task agents, we replace
the agent backbone with Qwen3-8B, Qwen3-32B, and DeepSeek-V4-Flash while
keeping the trained query generator and calibrated uncertainty
thresholds fixed. Table~\ref{tab:cross-agent} shows that our method improves
average success over both the base agent and the prompted Qwen3-8B
query generator for every backbone, reaching $28.17\%$, $40.25\%$, and
$48.20\%$, respectively. The gains persist as the task agent becomes stronger,
indicating that the learned uncertainty-resolution policy is not tied to the
behavior of a particular downstream agent. Appendix~\ref{app:routing-calibration}
separately evaluates GLM-4-32B as both the task agent and decomposer under
model-specific threshold recalibration.

\textbf{Cross-benchmark Generalization.}~
\label{sec:tau3-transfer}
To test generalization beyond the training environment, we apply the complete
pipeline to $\tau^3$-bench \citep{barres2025tau2} without retraining or threshold recalibration.
We evaluate on \textit{retail}, \textit{airline}, and \textit{telecom}, and restrict the study to transactional text domains, where banking and voice domains~\citep{shi2026tauknowledge,ray2026tauvoice} fall outside our current action space. Table~\ref{tab:tau3-results}
shows that our method achieves the highest average success ($29.00\%$) and the
fewest total steps (29.25), compared with $27.23\%$ average success for the
strongest baseline, VerifiAgent. This leading result indicates that the
learned query generator is not restricted to retail domain, but generalizes across interaction environment.

\subsection{Further Discussions}

\begin{wrapfigure}{r}{0.55\linewidth}
\vspace{-6mm}
\setlength{\abovecaptionskip}{2pt}   
\setlength{\belowcaptionskip}{0pt}
    \centering
    \includegraphics[width=\linewidth]{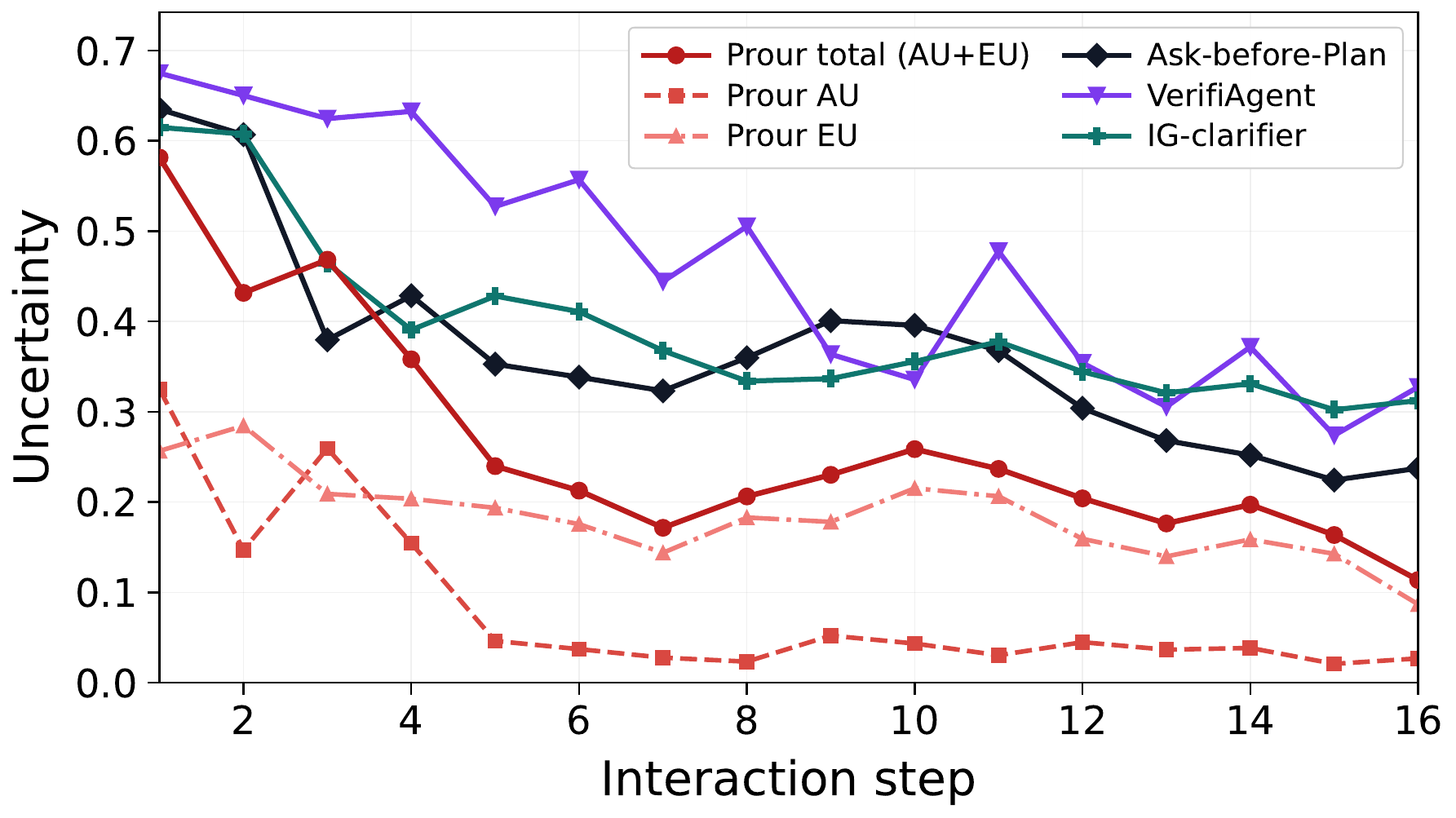}
    \caption{\textbf{Uncertainty trajectories over interaction.}
\textsc{Prour} reports AU, EU, and total uncertainty over the first 16 steps, while baselines report total uncertainty with decomposer applied post hoc to trajectories. }
    \label{fig:uncertainty_trajectory}
    \vspace{-3mm}
\end{wrapfigure}

\textbf{Uncertainty Resolution over Interaction.}~
To examine how uncertainty evolves during interaction,
Figure~\ref{fig:uncertainty_trajectory} reports uncertainty for \textsc{Prour} and baselines over the first 16 steps. \textsc{Prour}
reduces total uncertainty from $0.58$ to $0.11$ and maintains lower uncertainty throughout the trajectory.
AU decreases mostly in the early resolution of user-side ambiguity, whereas EU decreases gradually
from $0.26$ to $0.08$, reflecting continued acquisition
of world-side evidence. The small residual AU is expected because action distributions are estimated through sampling,
which preserves some variability even after the user intent is largely resolved.

\textbf{Performance--Cost Trade-off.}~
To quantify the trade-off between uncertainty decomposition and inference
cost, we conduct a post-hoc analysis varying the maximum number of decomposer calls. Figure~\ref{fig:decomposer-budget} shows that introducing a small budget initially reduces total interaction steps, suggesting that timely clarification and verification can prevent unnecessary downstream tool calls. With larger budgets, interaction steps rise to 20.42 at a budget of 16, while average success continues to improve from 18.37\% to 29.42\%. Beyond this point, larger budgets increase cost without further benefit and slightly degrade success, suggesting that repeated interventions accumulate redundant or weakly relevant evidence after major uncertainty has been resolved. The complete trade-off results and invocation-frequency analysis are reported in
Appendix~\ref{app:efficiency-details}.

\section{Conclusion}
\label{sec:conclusion}
We introduced \emph{uncertainty routing} for proactive tool-using agents, where uncertainty is resolved by acquiring information from the routed source. Our framework \textsc{Prour} decomposes action uncertainty into aleatoric uncertainty (AU) and epistemic uncertainty (EU), routing the agent among \textsc{act}, \textsc{clarify}, and \textsc{verify}. A query generator is trained with a mode-conditioned information-gain objective to acquire user-side information under \textsc{clarify} and grounded world-side evidence under \textsc{verify}. On $\tau$-bench, \textsc{Prour} improves
average success rate from $23.60\%$ to $28.17\%$ with fewer interaction steps, and generalizes without retraining to stronger task agents and to $\tau^3$-bench. These results support source-aligned uncertainty routing as a principled approach for proactive agents.

\section*{AI Use Statement}

We used generative AI tools to assist with drafting and editing parts of the manuscript for clarity, readability, and presentation. We also used generative AI tools to assist with annotating $\tau$-bench trajectories for the uncertainty case taxonomy in
Appendix~\ref{app:uncertainty_case_taxonomy}, and to generate initial uncertainty labels for the development-set routing analysis described in Appendix~\ref{app:routing-calibration}. All such annotations were manually reviewed and corrected by the authors before being used for calibration and analysis. We have not used generative AI tools to develop the conceptual framework, propose hypotheses, formulate mathematical claims, or implement the method. We have reviewed all AI-assisted work and take responsibility for the final content, analyses, claims, and results presented in this paper.

\bibliography{iclr2027_conference}
\bibliographystyle{iclr2027_conference}

\clearpage

\appendix
\section{Appendix}

\subsection{Uncertainty Case Taxonomy}
\label{app:uncertainty_case_taxonomy}
We construct the taxonomy from historical retail and airline $\tau$-bench
trajectories by treating every assistant response or tool call as a decision step. For each decision, GPT-5.5 receives the task instruction, recent
dialogue context, the current assistant decision, reference actions and
outputs, and any immediately following tool observations. It determines
whether uncertainty is present; classifies it as epistemic, aleatoric, or absent; and records the uncertain object, available evidence, resolution
outcome, and a provisional open-ended case label. The annotations are manually verified and corrected, to ensure that instructions
do not infer uncertainty solely from task failure or deviation from a reference
action, and include uncertainty that is handled successfully during an
interaction. We then normalize the annotations and consolidate the open labels
using the decision type, tool-call type, and described evidence, assigning each
retained decision step to one of the nine source-relevant cases in
Table~\ref{tab:tau_bench_uncertainty_cases}. The table reports 19,022
step-level events, rather than trajectory-level failures, so a trajectory can
contribute multiple events.

For source attribution, cases resolvable from profiles, tools, policy,
computation, or interaction history are categorized as world-side
(epistemic), whereas missing user-owned intent or preferences are categorized as user-side (aleatoric). Under this categorization,
16,947 of the 19,022 events (89.1\%) are world-side, while 2,075 (10.9\%) are
user-side (Figure~\ref{fig:uncertainty_cases_distribution}). Thus, clarification
alone addresses only a small portion of the observed uncertainty; most cases
require retrieving or verifying information about the world.

\begin{table}[h]
\centering
\small
\caption{
Uncertainty cases identified in $\tau$-bench trajectories; parentheses in
\textit{Examples} refer to representative decision steps in the trajectories.
}
\label {tab:tau_bench_uncertainty_cases}
\setlength{\tabcolsep}{3.5pt}
\renewcommand{\arraystretch}{1.12}
\begin{adjustbox}{max width=\textwidth}

\begin{tabular}{
    p{0.28\textwidth}
    c
    p{0.12\textwidth}
    p{0.5\textwidth}
}
\toprule
\multicolumn{1}{c}{\textbf{Cases}} &
\multicolumn{1}{c}{\textbf{Events}} &
\multicolumn{1}{c}{\textbf{Uncertainty}} &
\multicolumn{1}{c}{\textbf{Examples}} \\
\midrule

Entity grounding
& 6{,}438
& Epistemic
& Missing user ID and reservation ID; resolvable through profile lookup.
{\scriptsize(\texttt{gpt-4o-airline:0:decision:0})} \\

State or evidence lookup
& 3{,}880
& Epistemic
& Retrieve the current reservation state before acting.
{\scriptsize(\texttt{gpt-4o-airline:0:decision:3})} \\

Tool-argument verification
& 2{,}356
& Epistemic
& A return-flight search reverses the required origin and destination.
{\scriptsize(\texttt{gpt-4o-airline:10:decision:12})} \\

\makecell[l]{User intent\\ or preference clarification}
& 2{,}075
& Aleatoric
& The user must select a cancellation reason or return-flight option.
{\scriptsize(\texttt{gpt-4o-airline:10:decision:7})} \\

Calculation verification
& 1{,}751
& Epistemic
& The agent computes \$255, whereas tool evidence reports \$305.
{\scriptsize(\texttt{gpt-4o-airline:0:decision:12})} \\

Plan completion verification
& 1{,}243
& Epistemic
& The agent books a replacement without verifying that the original reservation was cancelled.
{\scriptsize(\texttt{gpt-4o-airline:10:decision:18})} \\

\makecell[l]{Unnecessary questions\\ answerable by retrieval}
& 645
& Epistemic
& Asks the user for an identifier or order detail available through tools.
{\scriptsize(\texttt{gpt-4o-airline:101:decision:7})} \\

\makecell[l]{Domain-rule\\ or policy retrieval}
& 596
& Epistemic
& Uncertain whether passenger removal or basic-economy exchange is permitted.
{\scriptsize(\texttt{gpt-4o-airline:10:decision:3})} \\

Partial completion
& 38
& Epistemic
& Declares success while a requested subgoal remains incomplete.
{\scriptsize(\texttt{gpt-4o-airline:100:decision:10})} \\

\bottomrule
\end{tabular}
\end{adjustbox}
\end{table}

\subsection{Experimental Details}
\label{app:experimental-details}

\subsubsection{Implementation Details}
\label{app:implementation-details}

\textbf{Interaction Trajectory Generation.}
We collect 500 trajectories from a $\tau$-bench environment augmented with a
query generator using the retail training split. The Qwen3-8B task agent interacts
with domain tools under the standard $\tau$-bench protocol, a Qwen3-32B teacher
query generator is invoked after tool calls to decide whether to ask a
follow-up query, and a Qwen3-8B user simulator provides the response. This
produces trajectories
$\tau=\{U,(a_t,o_t,Q_t,A_t)_{t=1}^{T},G\}$, where $U$ and $G$ are the initial message and task-defined goal, $(a_t,o_t)$ is a tool interaction, and
$(Q_t,A_t)$ is an optional query--response exchange. Importantly, the
teacher-generated queries used during data collection serve only to elicit
user feedback and enrich the interaction histories; we do not retain them as
SFT targets or convert them into preference pairs. Directly supervising on
these outputs would train the query generator to imitate the teacher rather
than optimize the mode-conditioned information-gain objective. Instead, all
query generator decisions and queries used for optimization are produced
by the Qwen3-1.7B policy during on-policy rollouts and scored by our reward.

\textbf{Step-level Instance Extraction.}
We split trajectories into disjoint training and validation partitions before extracting step-level instances. We then retain the first 16 agent decisions from each trajectory and construct the
history $x_t=(U,(a_i,o_i,Q_i,A_i)_{i<t})$. The calibrated decomposer assigns
each state to \textsc{act}, \textsc{clarify}, or \textsc{verify}; \textsc{act}
states are omitted because they do not invoke the query generator. This
yields 3,520 state--mode instances, divided into 2,816 training and 704
validation examples. For \textsc{clarify}, the reward target is the normalized
user goal $G^*$; for \textsc{verify}, it is the reference next action
$a_t^*$. As in the IG-Clarifier pipeline, goal normalization converts
third-person descriptions into first-person statements, rewrites structured
fields and key--value attributes as fluent natural-language descriptions, and
filters out metadata or other non-functional attributes that do not affect
the task. It preserves task-specific values, such as entities, quantities,
identifiers, preferences, and requested actions, so that the resulting goal
remains faithful to the original task while providing a consistent target for
training.

\textbf{Model and System Configuration.}
The task agent, decomposer, user simulator, and verifier are
frozen Qwen3-8B models with role-specific prompts. The query
generator is initialized from Qwen3-1.7B and is the only optimized component.
We train with DAPO under the VERL framework. We use FlashAttention-2 and
gradient checkpointing during training. Rollouts are generated with vLLM, with
chunked prefill enabled and GPU memory utilization set to $0.20$. The maximum
sequence length is 3,160 tokens, comprising at most 1,400 prompt tokens and
1,760 response tokens. Dialogue histories are truncated from the left when
they exceed this maximum length. Training runs on one node with two 80\,GiB
NVIDIA H100 GPUs.

\textbf{Optimization, Batching, and Sampling.}
We use the GRPO advantage estimator with asymmetric PPO clipping. The learning
rate is $7\times10^{-7}$ with 60 warmup steps; the lower and upper clipping
bounds are $0.20$ and $0.30$, and the entropy coefficient is $0.01$. Weight
decay is $0.01$, gradients are clipped to norm $1.0$, losses use token-level
mean reduction, and no explicit KL loss is applied. At each training step, 16
prompts each produce eight candidates, giving 128 rollouts; six candidates per
prompt are retained for optimization. Rollouts use temperature $0.8$,
top-$p=1.0$, and no top-$k$ truncation. We perform four PPO epochs with
mini-batches of one prompt and micro-batches of one per GPU. Training data are
shuffled, and Qwen3 thinking is enabled. We optimize for 2,010 steps, save
checkpoints every 100 steps, and use the final checkpoint for evaluation.

\textbf{Baselines and Ablations.}
We compare against the base agent without proactive intervention and five
representative information-seeking approaches. IG-Clarifier~\citep{deng2026uncertainty}
learns clarification questions and resolves uncertainty only by querying the
user. Ask-before-Plan~\citep{zhang2024askbeforeplan} predicts clarification needs before planning but does not explicitly decompose and route uncertainty. VerifiAgent~\citep{han2025verifiagent} performs meta- and tool-based
verification without clarifying missing
information from the user.  SABER~\citep{cuadron2025saber} gates explicit user confirmation on mutating actions, adds targeted reflection, and cleans context, so it requests user consent rather than acquiring world-side evidence.
SAGE-Agent~\citep{suri2026structured} tracks structured uncertainty over tool parameters and asks the highest expected-value-of-perfect-information question to the user.
They are all reimplemented on the Qwen3-8B backbone.
Each controlled ablation changes a single component: \textsc{LLM Router} replaces the AU/EU decomposer with a
prompted Qwen3-8B router; \textsc{Clarify-only} disables the \textsc{verify}
branch; and \textsc{Verify-only} disables the \textsc{clarify} branch. We also
replace the trained query generator with prompted Qwen3-8B, Qwen3-32B, or
DeepSeek-V4-Flash while keeping the remaining pipeline fixed.

\textbf{Evaluation Setting.}
We evaluate on 115 held-out retail tasks and 50 out-of-distribution airline
tasks. Unless otherwise stated, we use decoding temperatures of $0.01$, $1.0$, and $0.0$ for the task agent, user simulator,
and query generator, respectively. $\mathrm{Pass}@{1}$ is the percentage of
tasks receiving the benchmark's binary success reward averaged over three random seeds, and \textit{Average} is the unweighted mean across domains. For efficiency, we report wall-clock seconds and total interaction
steps partitioned into \textsc{Act}, \textsc{Clarify}, and \textsc{Verify} turns. For $\tau^3$-bench, we use release v1.0.1, which incorporates the maintainers' task corrections based on SABER~\citep{cuadron2025saber}.

\subsubsection{Uncertainty Estimation and Routing Calibration}
\label{app:routing-calibration}
{
\begin{figure}[t]
    \centering
    \begin{subfigure}[t]{0.495\textwidth}
        \centering
        \includegraphics[width=\linewidth]{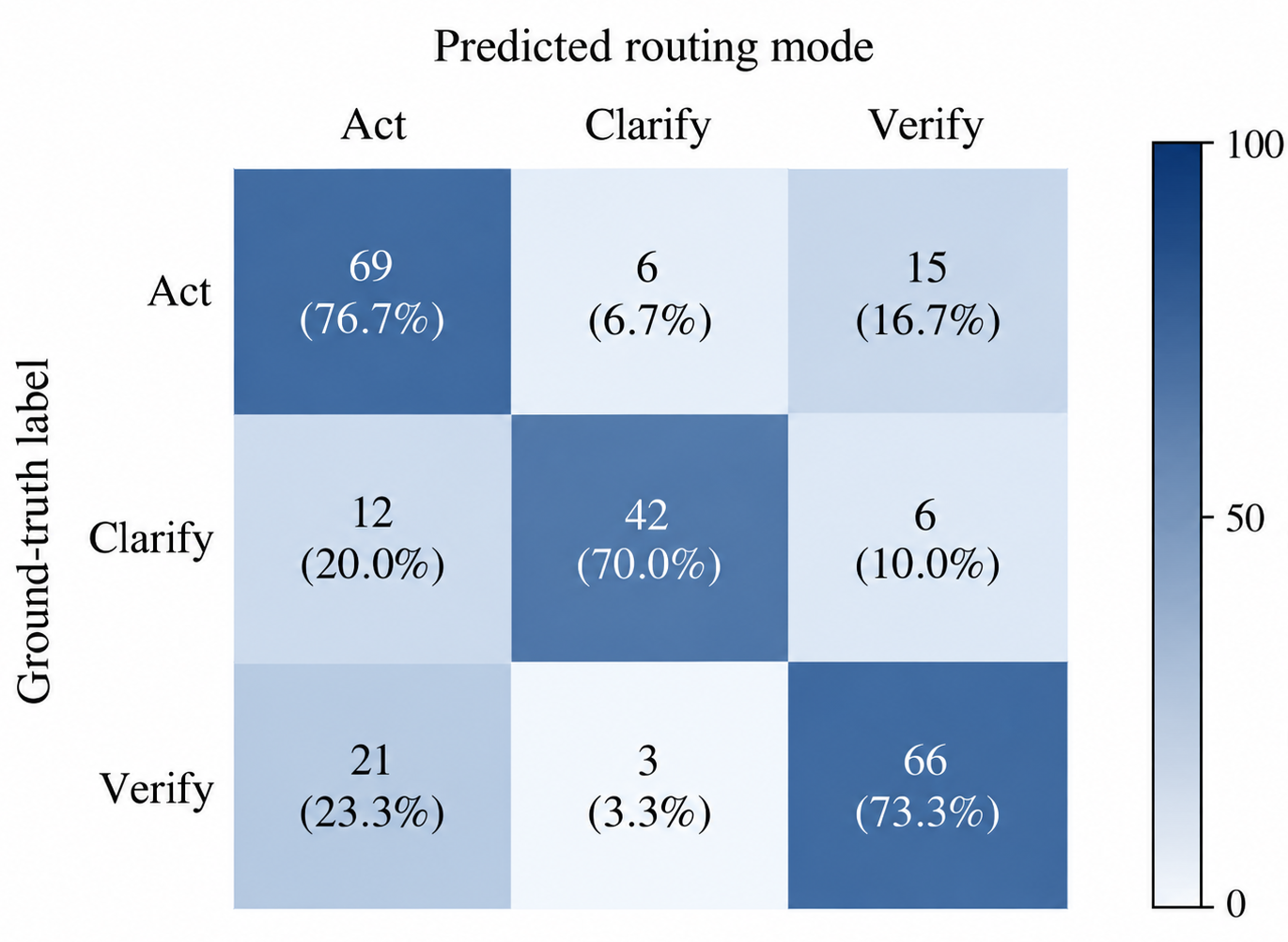}
        \caption{Calibrated \textsc{Prour} Router}
        \label{fig:decomposer_confusion_matrix}
    \end{subfigure}
    \begin{subfigure}[t]{0.495\textwidth}
        \centering
        \includegraphics[width=\linewidth]{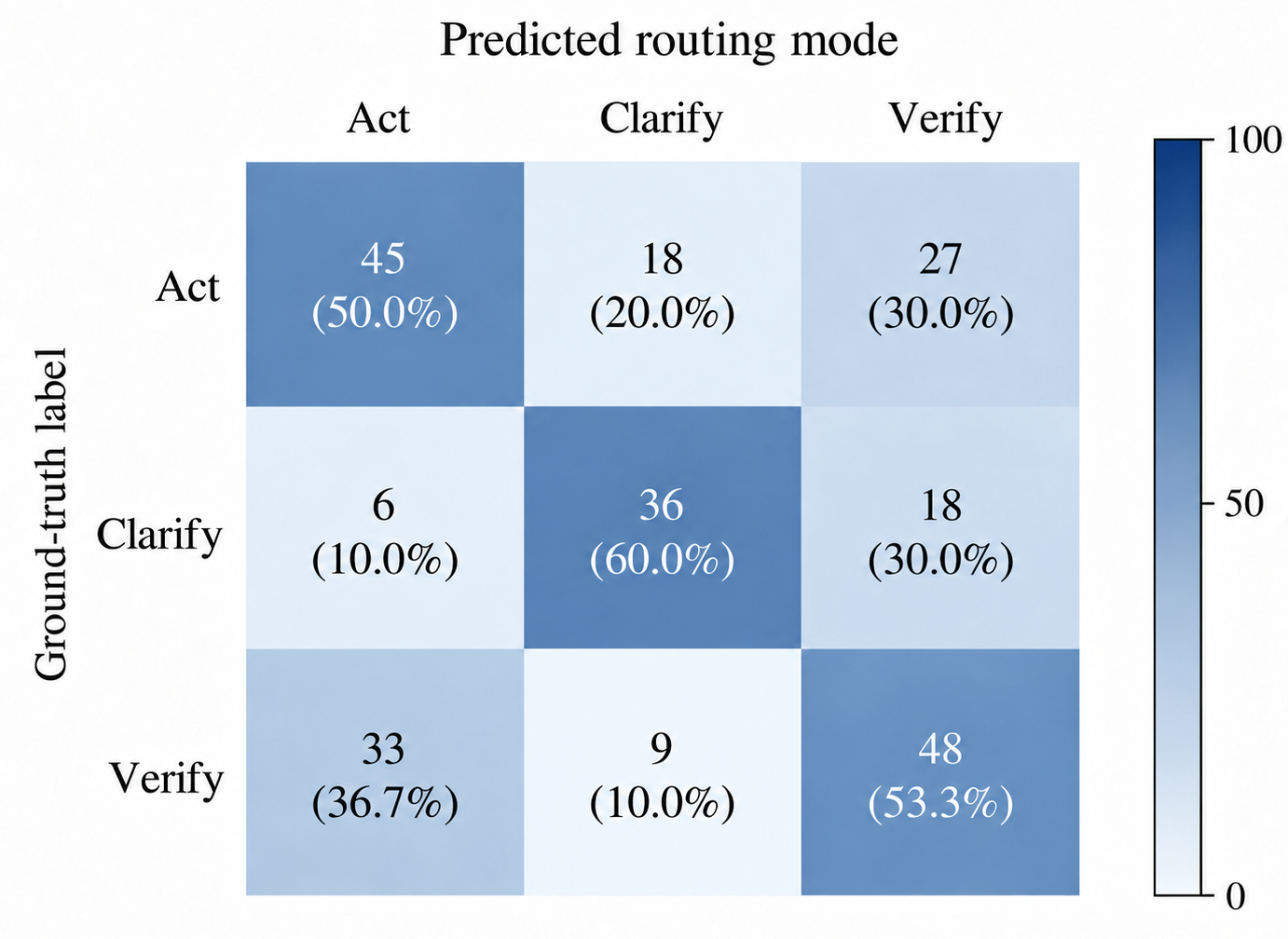}
        \caption{Prompted LLM Router}
        \label{fig:llm_confusion_matrix}
    \end{subfigure}
    \caption{\textbf{Held-out routing performance of our calibrated \textsc{Prour} and the prompted LLM Router.}
    Rows denote ground-truth labels and columns denote predicted modes;
    each cell reports the count and row-normalized percentage. \textsc{Prour}
    applies the Qwen3-8B thresholds selected on the retail development set,
    whereas the prompted Qwen3-8B router predicts a mode directly, without calibrated thresholds. Both methods are evaluated on the same
    disjoint airline test set.}
    \label{fig:routing-confusion}
\end{figure}
}
\textbf{Uncertainty is Model-specific.}
Because AU and EU are computed from a particular model's predictive
distributions, their numerical scales are model-specific~\citep{kendall2017uncertainties}. Language-model
confidence and knowledge boundaries vary across architectures and capabilities~\citep{jiang2021know,kadavath2022language}. Replacing the model used for goal
proposal and action scoring can therefore change both the action distributions and resulting uncertainty scores. We consequently recalibrate
$(\tau_{\mathrm{AU}},\tau_{\mathrm{EU}})$ when the agent and decomposer backbone changes, accounting for
model-dependent shifts in the AU and EU distributions.

\textbf{Uncertainty Estimation Protocol.}
At each decision step, the decomposer independently samples $K=5$ plausible
goal interpretations and, for each interpretation, independently samples
$L=4$ structured actions. A deterministic parser requires exactly the top-level fields \texttt{tool} and \texttt{arguments} and validates the action against the corresponding tool schema; invalid outputs are discarded without resampling. Valid actions are canonicalized and deduplicated across all sampled actions to form one step-specific candidate set shared across interpretations.  Each action is then scored under
each goal interpretation by summed teacher-forced token log-likelihood, including
the common completion token, and normalized over this candidate set to obtain the
distributions used for estimating AU and EU. Appendix~\ref{app:decomposer-prompts} provides the proposing, sampling, and scoring prompts. The procedure requires $K(L+2)$ model calls when all candidates are scored as one batch per interpretation (30 model calls for $K=5, L=4$).

\textbf{Calibration Protocol.}
To avoid test-set leakage, we calibrate routing thresholds on a disjoint
$\tau$-retail development set of 240 decision steps. Each step is first
assigned one of three uncertainty labels---AU, EU, or \textsc{No
Uncertainty}---using GPT-5.5 under the fixed annotation rubric in
Appendix~\ref{app:prompt-templates}; all labels are then manually reviewed
and corrected by the authors. These labels correspond directly to the
routing modes: AU to \textsc{clarify}, EU to \textsc{Verify}, and
\textsc{No Uncertainty} to \textsc{Act}. The resulting set contains
60 \textsc{clarify}, 90 \textsc{Verify}, and 90 \textsc{Act} decisions. For held-out routing evaluation, we independently sample and annotate a disjoint set of 240 airline decision steps using the same protocol. This test set is used only after threshold selection.

The decomposer estimates AU and EU based on Eq.~\ref{eq:au-score}--\ref{eq:eu-score}. We perform a grid search over
$(\tau_{\mathrm{AU}},\tau_{\mathrm{EU}})\in[0,0.8]^2$ with step size
$0.01$, applying the AU-first routing rule in Eq.~\ref{eq:routing}, and
select the threshold pair that maximizes three-class macro-F1 on the
development set. For the default Qwen3-8B decomposer, this yields
$(\tau_{\mathrm{AU}},\tau_{\mathrm{EU}})=(0.25,0.20)$ with a macro-F1
of 0.714.  We then freeze these thresholds and apply them unchanged to the
airline test set. As a threshold-free comparator, the prompted
Qwen3-8B LLM Router receives the same decision history and available
tools, but directly predicts \textsc{act}, \textsc{clarify}, or
\textsc{verify} without estimating AU or EU. As shown in
Figure~\ref{fig:routing-confusion}, our calibrated \textsc{Prour} router
achieves a macro-F1 of $74.04\%$, compared with $54.24\%$ for the prompted
LLM Router. This improvement supports the
benefit of explicit uncertainty decomposition with calibrated thresholds over direct prompted routing.

\begin{wraptable}{r}{0.50\textwidth}
\vspace{-4mm}
\centering
\small
\caption{\textbf{Routing F1 after model-specific calibration.}
Macro-F1 and per-class F1 evaluate three-way routing on the same development set.}
\label{tab:routing-calibration-summary}


\begin{tabular}{lcc}
\toprule
\textbf{Metric} & \textbf{Qwen3-8B} & \textbf{GLM-4-32B} \\
\midrule
$(\tau_{\mathrm{AU}},\tau_{\mathrm{EU}})$
& $(0.25,0.20)$ & $(0.20,0.08)$ \\
Macro-F1
& 0.714 & 0.711 \\
\textsc{act} F1
& 0.700 & 0.742 \\
\textsc{clarify} F1
& 0.683 & 0.706 \\
\textsc{verify} F1
& 0.760 & 0.685 \\
\bottomrule
\end{tabular}

\end{wraptable}

Because uncertainty scores are model-specific, we repeat the same
calibration procedure for GLM-4-32B, obtaining
$(\tau_{\mathrm{AU}},\tau_{\mathrm{EU}})=(0.20,0.08)$ with a macro-F1
of 0.711. All thresholds are selected before test evaluation and remain
fixed thereafter. Table~\ref{tab:routing-calibration-summary}
summarizes the corresponding class-level F1 scores. The two backbones achieve
similar macro-F1 but differ by mode: GLM-4-32B obtains higher F1 on
\textsc{act} and \textsc{clarify}, whereas Qwen3-8B obtains higher F1 on
\textsc{verify}.

Table~\ref{tab:cross-agent-recalibration} tests whether this recalibration affects downstream performance while fixing GLM-4-32B as both the task agent and decomposer. Model-matched thresholds increase average success from
$33.26\%$ to $37.00\%$ with our trained query generator and from $29.65\%$
to $31.95\%$ with the prompted Qwen3-8B generator. The improvement after model-matched recalibration is consistent across both query generators, providing empirical evidence that uncertainty distribution is model-specific. For our generator, higher success is accompanied by more intervention steps, indicating that recalibration changes
the routing policy's intervention rate rather than providing a cost-free gain.

\begin{table}[t]
\centering
\small
\caption{
\textbf{Effect of model-matched routing thresholds.} GLM-4-32B serves as both
the task agent and decomposer in all rows. The upper three rows use thresholds calibrated for Qwen3-8B, and the lower three rows use thresholds calibrated for GLM-4-32B. Query generators are the same across the two blocks. The \textit{None} rows are base agent without decomposer, and do not use routing thresholds.
}
\label {tab:cross-agent-recalibration}
\setlength{\tabcolsep}{3.5pt}
\renewcommand{\arraystretch}{1.12}
\begin{adjustbox}{max width=\textwidth}
\begin{tabular}{llccccccc}
\toprule
\multicolumn{2}{c}{\textbf{Setting}} &
\multicolumn{3}{c}{\textbf{Success Rate (\% $\uparrow$)}} &
\multicolumn{4}{c}{\textbf{Steps $\downarrow$}} \\
\cmidrule(lr){1-2}
\cmidrule(lr){3-5}
\cmidrule(lr){6-9}

\textbf{Calibrated Model} &
\textbf{Query Generator} &
\textbf{Retail} &
\textbf{Airline} &
\textbf{Average} &
\textbf{Avg. \textsc{act}} &
\textbf{Avg. \textsc{clarify}} &
\textbf{Avg. \textsc{verify}} &
\textbf{Total} \\
\midrule

\multirow{3}{*}{Qwen3-8B}
& None
& 18.26 & 14.00 & 16.13
& 31.44 & -- & -- & 31.44 \\

& Qwen3-8B
& 31.30 & 28.00 & 29.65
& 22.72 & 2.41 & 14.12 & 39.25 \\

\rowcolor{blue!10}
& \textsc{Prour}
& \textbf{36.52}
& \textbf{30.00}
& \textbf{33.26}
& 19.13
& 1.24
& 10.46
& \textbf{30.83} \\
\midrule

\multirow{3}{*}{GLM-4-32B}
& None
& 18.26 & 14.00 & 16.13
& 31.44 & -- & -- & \textbf{31.44} \\

& Qwen3-8B
& 33.90 & 30.00 & 31.95
& 21.17 & 1.27 & 15.65 & 38.09 \\

\rowcolor{blue!10}
& \textsc{Prour}
& \textbf{40.00}
& \textbf{34.00}
& \textbf{37.00}
& 20.62
& 2.67
& 11.69
& 34.98 \\

\bottomrule
\end{tabular}
\end{adjustbox}
\vspace{-3mm}
\end{table}

\subsubsection{Performance--Cost Trade-off}
\label{app:efficiency-details}

Table~\ref{tab:decomposer-budget-frequency} reports the complete trade-off evaluation under two settings. The budget limits the maximum number of decomposer calls per episode. When the budget is exhausted, the conversation continues between the user and base agent, without decomposer or query generator. The frequency setting invokes the decomposer once every $n$ agent decision steps under an unlimited budget.

\begin{table}[t]
\centering
\small
\caption{
Performance on $\tau$-bench under different decomposer-call budgets and
invocation frequencies. \textit{Total steps} counts the number of interaction steps.}
\label {tab:decomposer-budget-frequency}
\setlength{\tabcolsep}{3.5pt}
\renewcommand{\arraystretch}{1.12}
\begin{adjustbox}{max width=\textwidth}
\begin{tabular}{lcccccc}
\toprule
\multicolumn{2}{c}{\textbf{Setting}} &
\multicolumn{3}{c}{\textbf{Success Rate (\% $\uparrow$)}} &
\multicolumn{2}{c}{\textbf{Cost $\downarrow$}} \\
\cmidrule(lr){3-5}
\cmidrule(lr){6-7}

\textbf{} &
\textbf{} &
\textbf{Retail} &
\textbf{Airline} &
\textbf{Average} &
\textbf{Total steps} &
\textbf{Time/episode (s)} \\
\midrule

\multirow{8}{*}{Budget}
& 1 & 17.40 & 19.33 & 18.37 & 22.13 & 515.45 \\
& 2 & 19.42 & 24.00 & 21.71 & 16.26 & 458.49 \\
& 4 & 23.50 & 28.00 & 25.75 & 14.86 & 481.20 \\
& 8 & 24.30 & 30.00 & 27.15 & 16.36 & 538.01 \\
& 16 & 27.50 & 31.33 & 29.42 & 20.42 & 559.16 \\
& 32 & 26.38 & 30.00 & 28.19 & 21.65 & 669.24 \\
& Unlimited & 27.00 & 29.33 & 28.17 & 21.33 & 696.51 \\
\midrule

\multirow{4}{*}{Frequency}
& Every 1 & 27.00 & 29.33 & 28.17 & 21.33 & 696.51 \\
& Every 2 & 24.64 & 30.00 & 27.32 & 25.26 & 708.27 \\
& Every 4 & 20.00 & 26.00 & 23.00 & 23.77 & 432.93 \\
& Every 8 & 19.10 & 20.00 & 19.55 & 22.18 & 400.56 \\

\bottomrule
\end{tabular}
\end{adjustbox}
\end{table}

\begin{figure}[t]
    \centering
    \begin{subfigure}[t]{0.495\textwidth}
        \centering
        \includegraphics[width=\linewidth]{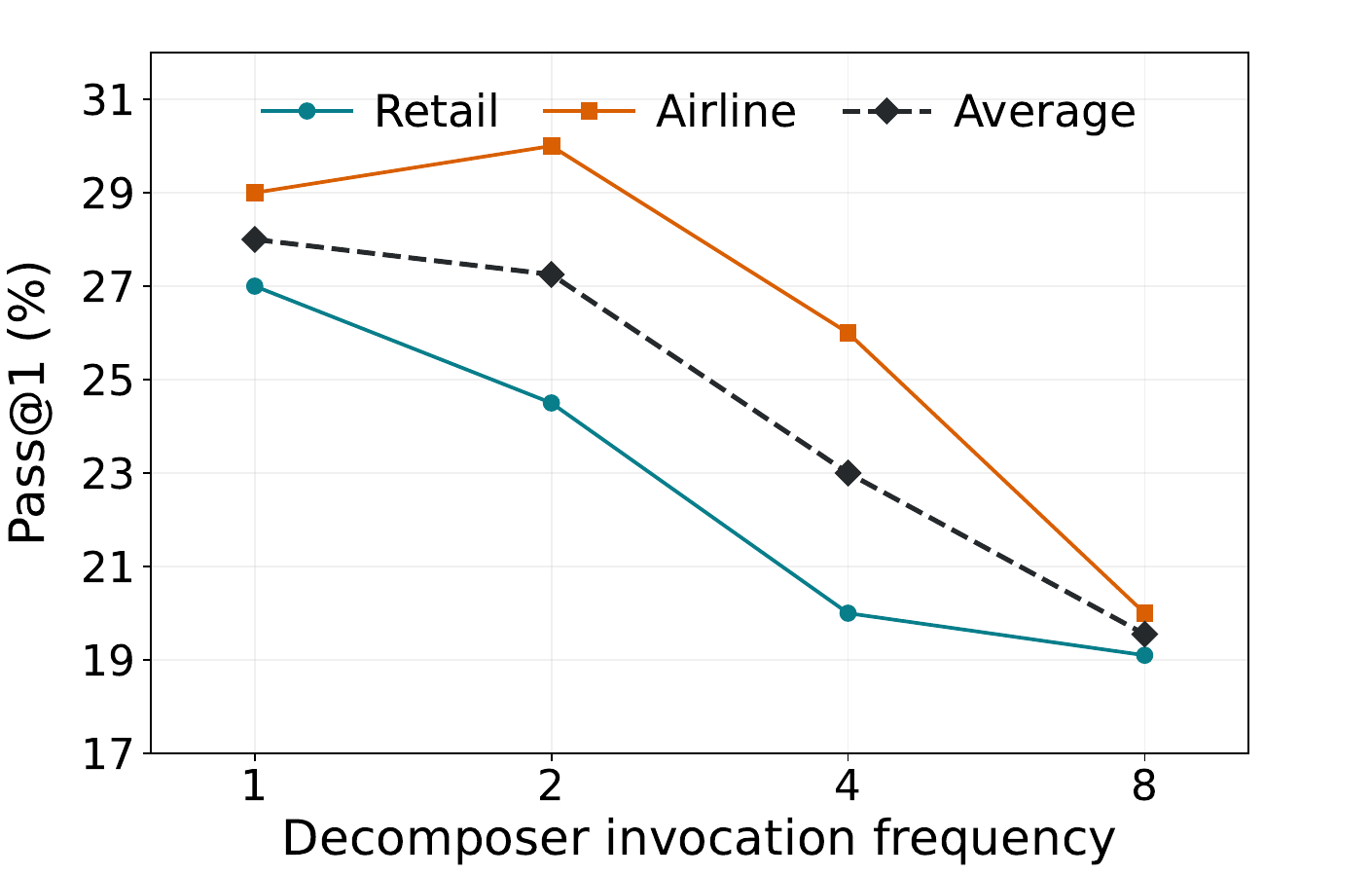}
        \caption{Task success w.r.t. invocation frequencies.}
        \label{fig:decomposer-frequency-success}
    \end{subfigure}
    \begin{subfigure}[t]{0.495\textwidth}
        \centering
        \includegraphics[width=\linewidth]{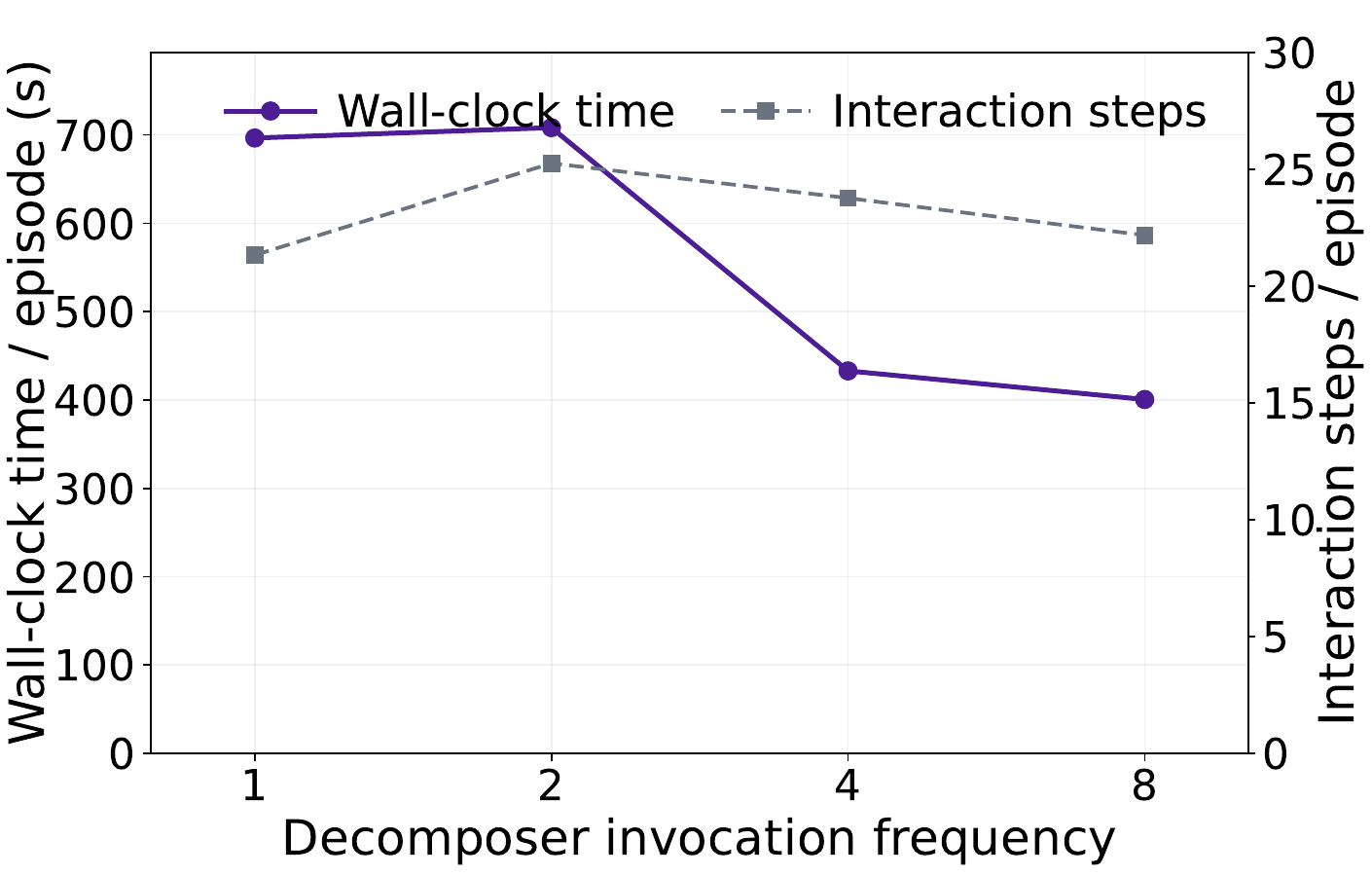}
        \caption{Average wall-clock time and interaction steps.}
        \label{fig:decomposer-frequency-cost}
    \end{subfigure}

    \caption{\textbf{Effect of decomposer invocation frequency on performance
    and cost.} (a) $\mathrm{pass}$@1 on retail, airline, and their average.
    (b) Average wall-clock time and total interaction steps per episode.
    Each x-axis value $n$ denotes invoking the decomposer once every $n$ agent decision steps.}
    \label{fig:decomposer-frequency}
\end{figure}

As shown in Figure~\ref{fig:decomposer-frequency}, reducing the frequency from every step to every second step preserves most of the success rate ($27.32\%$ versus $28.17\%$), but increases both
interaction steps ($25.26$ versus $21.33$) and wall-clock time ($708.27$ versus
$696.51$ seconds). Skipping a timely routing opportunity can therefore prolong the trajectory by forcing the agent to compensate with additional steps, which can outweigh the compute saved by fewer decomposer calls.
Invoking the decomposer every four or eight decisions lowers interaction steps and wall-clock time, but average success drops sharply to $23.00\%$ and $19.55\%$. The trade-off is therefore non-monotonic: moderately sparse routing can lengthen an episode enough to erase its savings by fewer decomposer calls, whereas very sparse routing lowers runtime only at the cost of missing timely clarification and
verification, which substantially degrades task success. Among the evaluated frequency settings, decomposing and routing every step provides the best overall balance of task success and interaction efficiency.

\subsection{Mode-Conditioned Information-Gain Training Framework}
\label{app:dapo-details}

We generalize the belief-update estimator and group-relative DAPO optimization of
\citet{deng2026uncertainty} to the mode-conditioned source and
target in Equation~\ref{eq:mode-conditioned-source-target}. This appendix
provides the training details omitted from the main text.

\textbf{Belief Update.} The query generator serves as two roles during
training.  It generates candidate queries through $q_{\theta}$. During reward computation, it also serves in teacher-forcing mode as belief scorer to evaluate its own belief in the correct target through $b_{\theta}$, with gradients
detached through the scoring operation. For a target sequence
$Y_t=(y_1,\ldots,y_L)$ and context $c$, we use the length-normalized
log-likelihood
\begin{equation}
    \ell_{\theta}(Y_t\mid c)
    =\frac{1}{L}\sum_{j=1}^{L}
      \log b_{\theta}(y_j\mid c,y_{<j}).
    \label{eq:normalized-log-likelihood}
\end{equation}
The prior and posterior contexts are $(h_t,m_t)$ and
$(h_t,m_t,Q_t,A_t)$, respectively, which yields the reward in
Equation~\ref{eq:mode-conditioned-ig-reward}. Length normalization makes the
reward comparable across targets of different lengths. If the query
generator emits \texttt{NO}, the posterior equals the prior and we set $R_t=0$. 

We intentionally use the current query generator as the belief scorer so that information gain is measured in the policy's own evolving decision space: reward measures whether newly acquired evidence makes the target more identifiable to the policy being optimized, rather than to a separate frozen model with potentially different knowledge or calibration.

\textbf{Mode-Conditioned Rollouts.} For each state--mode pair $(h_t,m_t)$, the
rollout policy samples a group
$\{Q_t^{(1)},\ldots,Q_t^{(M)}\}$.

To prevent the query generator from hacking reward by eliciting informative responses, in \texttt{clarify} mode, a strict user simulator conditioned on $G^*$ responds to each candidate. It reveals
 task-specific information only when the query is precise and relevant, so high rewards are assigned to queries that truly require clarification. In
\texttt{verify} mode, the frozen verifier conditions on the domain policy,
interaction history, and permitted read-only tool evidence. It may retrieve
state, check tool arguments and policy constraints, recompute values, or assess
subgoal completion, but returns \texttt{insufficient evidence} when no grounded response is available. Neither responder receives gradient. Their prompt templates are provided in Appendix~\ref{app:prompt-templates}.

For each response $A_t^{(i)}$, Equation~\ref{eq:mode-conditioned-ig-reward}
produces reward $R_t^{(i)}$. The empirical mean
\begin{equation}
    \widehat{\mathcal U}_t
    =\frac{1}{M}\sum_{i=1}^{M}R_t^{(i)}
    \label{eq:monte-carlo-utility}
\end{equation}
is the Monte Carlo estimate of the expected belief-update utility during
on-policy sampling.

\textbf{Group-Relative Advantage.} To reduce reward variance, candidates from
the same state--mode pair are normalized within the rollout group:
\begin{equation}
    \widehat A_i
    =\frac{R_t^{(i)}-\mu_M(R_t)}{\sigma_M(R_t)+\epsilon},
    \label{eq:group-relative-advantage}
\end{equation}
where $\mu_M$ and $\sigma_M$ are the group mean and standard deviation and
$\epsilon$ is a numerical-stability constant. This group-relative normalization further reduces sensitivity to common shifts in the belief scorer's likelihood scale. Because the reward is defined for
the complete query--response exchange, the scalar advantage is assigned to all
generated tokens of candidate $i$.

\textbf{DAPO Objective.} Let $w_{i,j}$ be token $j$ in candidate $i$. The
importance ratio between the current and rollout policies is
\begin{equation}
    \rho_{i,j}(\theta)
    =\frac{q_{\theta}(w_{i,j}\mid h_t,m_t,w_{i,<j})}
           {q_{\theta_{\mathrm{old}}}(w_{i,j}\mid h_t,m_t,w_{i,<j})}.
    \label{eq:dapo-importance-ratio}
\end{equation}
Following DAPO \citep{yu2025dapo}, we optimize the token-normalized clipped
objective
\begin{equation}
    \mathcal{J}(\theta)
    =\frac{1}{\sum_i |Q_t^{(i)}|}
      \sum_{i=1}^{M}\sum_{j=1}^{|Q_t^{(i)}|}
      \min\!\left(
        \rho_{i,j}\widehat A_i,
        \operatorname{clip}
        \bigl(\rho_{i,j},1-\epsilon_{\mathrm{low}},
              1+\epsilon_{\mathrm{high}}\bigr)\widehat A_i
      \right),
    \label{eq:dapo-objective}
\end{equation}
where $\epsilon_{\mathrm{high}}>\epsilon_{\mathrm{low}}$ gives favorable
tokens a larger exploration range than unfavorable tokens.  Every rollout group shares the same history, routing mode, response source, and target type.  As training progresses, $q_{\theta}$ increasingly concentrates probability mass on queries that yield high information gain.

\subsection{Prompt Templates}
\label{app:prompt-templates}

\subsubsection{Uncertainty Decomposer Prompt Templates}
\label{app:decomposer-prompts}
The decomposer uses three prompts. The first prompt samples one plausible user-goal
interpretation. Conditioned on that interpretation, each independent call to
the second prompt samples one structured next-action candidate. Each canonical action is separately scored under the third prompt via teacher-forcing.

\begin{tcolorbox}[breakable, colback=gray!20, colframe=gray!50, title={Goal-Interpretation Proposing Prompt}, fontupper=\ttfamily\footnotesize]
You are analyzing a dialogue between a $\tau$-bench customer-service agent and a user. Your job is to infer one plausible interpretation of what the user is ultimately trying to accomplish, given the
dialogue so far.
\par\smallskip
Rules:\\
- Write exactly one plausible interpretation of the user's underlying goal.\\
- Write the interpretation in the first person and as one concise sentence.\\
- Infer the goal from the dialogue history without assuming information that has not been provided.\\
- Do not merely repeat the surface phrasing of the dialogue; capture the underlying goal.\\
- When the dialogue is ambiguous, output one plausible interpretation rather than enumerating multiple alternatives.\\
- Output only the interpretation and nothing else.
\par\smallskip
[Dialogue history]\\
\{history\}
\par\smallskip
[Output --- one plausible goal interpretation]
\end{tcolorbox}

\begin{tcolorbox}[breakable, colback=gray!20, colframe=gray!50, title={Candidate-Action Sampling Prompt}, fontupper=\ttfamily\footnotesize]
You are a $\tau$-bench customer-service agent. Sample exactly one plausible
next action using the goal interpretation, dialogue history, and available tool
schemas below.
\par\smallskip
[Goal interpretation]\newline
\{goal\}
\par\smallskip
[Dialogue history]\newline
\{history\}
\par\smallskip
[Available tools and argument schemas]\newline
\{tools\_with\_schemas\}
\par\smallskip
Rules:\newline
- Return exactly one action, not a list of alternatives.\newline
- For an environment-tool action, use one available tool name and provide a
complete arguments object conforming to that tool's schema.\newline
- Preserve argument values exactly when they occur in the goal interpretation
or dialogue. Do not invent identifiers, dates, quantities, or other unsupported
facts.\newline
- If the appropriate next action is to answer the user without an environment
tool, use the special tool name ``respond'' with an empty arguments object.\newline
- Output strict JSON only, without Markdown or explanation.
\par\smallskip
Output format:\newline
\{"tool": "<tool\_name>", "arguments": \{...\}\}
\end{tcolorbox}

\begin{tcolorbox}[breakable, colback=gray!20, colframe=gray!50, title={Teacher-Forced Action Scoring Prompt}, fontupper=\ttfamily\footnotesize]
You are a $\tau$-bench customer-service agent. Determine the most plausible
next action from the goal interpretation, dialogue history, and available tool
schemas. Represent the next action as exactly one strict JSON object with the
top-level fields ``tool'' and ``arguments.''
\par\smallskip
[Goal interpretation]\newline
\{goal\}
\par\smallskip
[Dialogue history]\newline
\{history\}
\par\smallskip
[Available tools and argument schemas]\newline
\{tools\_with\_schemas\}
\par\smallskip
[Next action --- canonical JSON target follows]
\end{tcolorbox}

\subsubsection{Query Generator Prompt Templates}
The query generator uses the \textsc{clarify} or \textsc{verify} template conditioned on the routed mode. The same templates are used by prompted Query Generator comparisons, with only the
underlying model changed.

\begin{tcolorbox}[breakable, colback=gray!20, colframe=gray!50, title={Query Generator Prompt for Clarify Mode}, fontupper=\ttfamily\footnotesize]
Role\newline
You are a Query Assistant responsible for determining whether the agent needs to ask a clarifying question.
\par\smallskip
History:\newline
\{history\}
\par\smallskip
Instructions:\newline
Analyze the dialogue and decide whether clarification is needed.\newline
Any reasoning, if you use it, must be enclosed between <think> and </think>.\newline
The final answer must come after this complete reasoning block.\newline
If one short question is needed before acting, reply exactly:\newline
YES [QUESTION]your question here[/QUESTION]\newline
If no question is needed, reply exactly:\newline
NO\newline
Ask only one concise question about user intent.\newline
Do not ask about data retrievable from tools.\newline
Do not repeat previous questions.\newline
Answers written only inside <think> are invalid.
\par\smallskip
expected output:\newline
Either:\newline
- NO\newline
or\newline
- YES [QUESTION]...[/QUESTION]
\end{tcolorbox}

\begin{tcolorbox}[breakable, colback=gray!20, colframe=gray!50, title={Query Generator Prompt for Verify Mode}, fontupper=\ttfamily\footnotesize]
You are a Query Assistant operating in VERIFY mode. The agent is uncertain about a fact in the world (e.g., the value of an order field, a policy rule, the result of a computation). Your job is to decide whether to issue a verification query that will be answered by a separate Verifier module with tool access, and if so, to phrase the precise query.
\par\smallskip
Output format (strict):\newline
Any reasoning, if you use it, must be enclosed between <think> and </think>.\newline
The final answer must come after this complete reasoning block.\newline
If a verification query is needed, reply exactly:\newline
YES [QUESTION]your question here[/QUESTION]\newline
If no verification is needed, reply exactly:\newline
NO
\par\smallskip
Guidelines:\newline
- The query must be specific and answerable by retrieving evidence (tool output, computation, policy snippet).\newline
- Examples of good queries: "What items are in order \#W7449508?", "Is basic-economy exchangeable under current policy?", "What is the total fare for HAT136 + HAT039 in economy?"\newline
- Do NOT ask the user --- the user already provided their intent. Ask about world state.\newline
- One query only. Keep it under 30 words.\newline
Answers written only inside <think> are invalid.
\par\smallskip
[Dialogue history]\newline
\{history\}
\par\smallskip
[Your decision]
\end{tcolorbox}

The belief scorer (same model as query generator) uses prompts to estimate prior and posterior belief before and after the query--response exchange. For the prior score, the mode-conditioned turn placeholder is empty. In \textsc{clarify} mode it predicts the
normalized user goal; in \textsc{verify} mode it predicts the reference next
tool action.

\begin{tcolorbox}[breakable, colback=gray!20, colframe=gray!50, title={Clarify-Mode Posterior Belief Prompt}, fontupper=\ttfamily\footnotesize]
You are an expert analyst. Summarize the user's hidden profile and intent based on the dialogue.
\par\smallskip
[Dialogue]\newline
\{history\}\newline
\{clarification\_turn\}
\par\smallskip
\#\#\# User Intent Summary:
\end{tcolorbox}

\begin{tcolorbox}[breakable, colback=gray!20, colframe=gray!50, title={Verify-Mode Posterior Belief Prompt}, fontupper=\ttfamily\footnotesize]
You are a tau-bench agent. Given the dialogue history and any verifier evidence, predict the next tool call the agent should make.
\par\smallskip
[Dialogue]\newline
\{history\}\newline
\{verification\_turn\}
\par\smallskip
\#\#\# Next tool call:
\end{tcolorbox}

\subsubsection{User Simulator and Verifier Prompt Templates}

The verifier receives only read-only tools together with the domain policy and
visible history. Its response must cite retrieved, computed, or policy evidence;
otherwise it returns \texttt{insufficient evidence}.

\begin{tcolorbox}[breakable, colback=gray!20, colframe=gray!50, title={Verifier Prompt}, fontupper=\ttfamily\footnotesize]
You are a Verifier. Your job is to answer a verification query about the current dialogue state by retrieving grounded evidence --- never by speculation.
\par\smallskip
You have access to a set of READ-ONLY tools (provided in the tools list). You may also reason over the dialogue history.
\par\smallskip
STRICT RULES:\newline
1. Your final answer MUST include at least one of:\newline
   (a) A citation of a tool you called and the value it returned (e.g., "get\_order\_details(order\_id=W7449508) returned: \{\{status: 'pending', ...\}\}").\newline
   (b) A computed value with the inputs shown (e.g., "Total = \$150 + \$155 = \$305 from get\_flight\_details on HAT136 and HAT039").\newline
   (c) A direct quote of a policy snippet from the system prompt or wiki.\newline
2. If you cannot reach a grounded answer (the needed tool is not available, or the data is missing), reply with exactly:\newline
     insufficient evidence\newline
3. Do NOT speculate. Do NOT invent values.\newline
4. Do NOT call any tool that mutates state (the tool list provided is already filtered to read-only).\newline
5. After at most \{max\_steps\} tool-call rounds, return your final natural-language answer.
\par\smallskip
[Domain policy / wiki]\newline
\{policy\}
\par\smallskip
[Dialogue history so far]\newline
\{history\}
\par\smallskip
[Verification query]\newline
\{question\}
\end{tcolorbox}

During training, the strict user simulator releases information only in response to a specific, relevant query.

\begin{tcolorbox}[breakable, colback=gray!20, colframe=gray!50, title={Strict User-Simulator Prompt for DAPO Training}, fontupper=\ttfamily\footnotesize]
Role\newline
You are a user interacting with an agent. Your behavior simulates a human user following a hidden instruction.
\par\smallskip
Your hidden instruction:\newline
\{G*\}
\par\smallskip
Instructions:\newline
conversation guidelines:\newline
- Respond with one message at a time using first-person statements.\newline
- Do not invent details that are not in the instruction; if something is unknown, say you do not remember it.\newline
- Rephrase the instruction in your own words and maintain a natural, human-like tone.
\par\smallskip
IMPORTANT - Handling vague or generic questions:\newline
- If the agent asks a vague, overly broad, or generic question (e.g., "Is there any additional information?", "Can you tell me more?", "Anything else?"), reply with: "No, that's all." or "Just do what I asked."\newline
- If the agent outputs a placeholder or acts out of character (e.g., "Your concise and specific clarifying question to the user here"), reply with: "Who are you talking to?"\newline
- For such questions, only provide information if it is DIRECTLY relevant to the current step of the instruction.\newline
- If the question is too vague to answer meaningfully, respond with "I'm not sure what specific information you need" or "Could you be more specific?"\newline
- Prefer specific, targeted questions that help the agent understand your exact needs.
\par\smallskip
Question: \{question\}
\par\smallskip
expected output:\newline
A single user utterance that follows the instruction and guidelines.
\end{tcolorbox}

During evaluation, user responses follow the standard $\tau$-bench simulator prompt instead of reusing the strict user prompt for training.

\begin{tcolorbox}[breakable, colback=gray!20, colframe=gray!50, title={User Simulator Prompt for Main Evaluation Results}, fontupper=\ttfamily\footnotesize]
You are a user interacting with an agent.
\par\smallskip
Instruction: \{instruction\}
\par\smallskip
Rules:\newline
- Generate one line at a time to simulate the user's message.\newline
- Do not reveal the entire instruction at once. Provide only information necessary for the current step.\newline
- Do not invent information absent from the instruction. If asked for an unavailable detail, say that you do not remember or have it.\newline
- If the instruction goal is satisfied, output \#\#\#STOP\#\#\# as a standalone message.\newline
- Do not repeat the instruction verbatim; express it naturally in your own words.\newline
- Maintain a natural conversation style and follow any personality specified in the instruction.
\end{tcolorbox}

The task agent itself follows the standard $\tau$-bench setup: the complete
domain policy is supplied as its system prompt and the environment exposes the
domain tool schemas through the model's tool interface. We add no task-specific
demonstration to the Qwen3-8B agent.

\subsubsection{Data Preparation and Calibration Prompt Templates}

The raw task instruction is converted into the normalized goal $G^*$ using the
following template.

\begin{tcolorbox}[breakable, colback=gray!20, colframe=gray!50, title={Goal-Normalization Prompt}, fontupper=\ttfamily\footnotesize]
Role\newline
You are a data converter. Your goal is to rewrite raw user detailed data into a flat, natural language format optimized for training smaller language models.
\par\smallskip
Transformation Rules:\newline
1. First-person normalization: convert third-person identifiers to first-person intent (e.g., "Your name is" to "My name is").\newline
2. Remove Noise: Remove all personality descriptions (e.g., "logical, shy, organized").\newline
3. Flatten JSON (Crucial): Remove all JSON syntax (\{\{,\}\},',:). Convert key-value pairs into natural phrases.\newline
   Use connectors like "with", "and", "set to".\newline
   Example: \{\{'color': 'red', 'size': 'M'\}\} $\rightarrow$ with color red and size M.\newline
4. Handle "(same as ...)": Keep the reference intact.\newline
   Ensure it flows logically with the details.\newline
   Example: \{\{'zip': '10228'\}\} (same as \#W123) $\rightarrow$ with zip 10228, which is the same as \#W123.\newline
5. Preserve Content: Do NOT delete specific values (like "144 Lakeview Drive"). Keep them just in case the user mentions them.
\par\smallskip
\#\#\# Example:\newline
Input:\newline
"Your name is Alex. You are shy and loud. For \#W1, change address to \{\{'street': '123 Main St', 'zip': '90210'\}\} (same as \#W2)."\newline
Output:\newline
"My name is Alex. For order \#W1, change address to street 123 Main St and zip 90210, which is the same as \#W2."
\par\smallskip
Input:\newline
"\{instruction\}"\newline
Output:
\end{tcolorbox}

In the development set, the labels for calibrating routing thresholds are annotated by GPT-5.5 with
the following fixed rubric before grid search.

\begin{tcolorbox}[breakable, colback=gray!20, colframe=gray!50, title={Uncertainty Labelling Prompt for Calibration}, fontupper=\ttfamily\footnotesize]
You are an expert annotator labeling agent uncertainty in $\tau$-bench dialogues.
\par\smallskip
At the decision point below, classify the next assistant move using EXACTLY
ONE of these labels:
\par\smallskip
- epistemic: the next assistant move should resolve a concrete fact, policy,
  or world state by looking something up, reading the wiki, or using a tool.\newline
- aleatoric: the next assistant move should ask the user because the missing
  information is a preference, choice, or selection among options.\newline
- no\_uncertainty: the agent already has enough information to act decisively
  right now.
\par\smallskip
Decision rule:\newline
- Choose epistemic for missing factual information that the assistant should
  look up or verify, including account/order/product identifiers, policy
  checks, and other concrete facts.\newline
- Choose aleatoric when the unresolved issue is a user preference, choice, or selection among valid options.\newline
- Do not use aleatoric for factual lookup requests such as order IDs, email addresses, zip codes, account numbers, product IDs, tracking numbers, or policy lookups.\newline
- Do not use aleatoric for identity verification or account recovery prompts such as "Could you share your order number?", "What is your email address?"
  Those are epistemic because the assistant is asking for a concrete fact.\newline
- Use aleatoric when the user must express a genuine preference or pick 
  between valid alternatives, such as color, size, shipping speed, refund
  destination, or which of several acceptable options to take.\newline
- If the assistant can directly call a tool or finish the request without
  asking anything else, choose no\_uncertainty.
\par\smallskip
Examples:\newline
- The agent needs an order ID, account number, product ID, or policy rule from
  the wiki:\newline
  epistemic\newline
- The agent needs the user's order number, email, zip code, account name, or tracking number to continue:\newline
  epistemic\newline
- The user has to choose between two colors, two shipping speeds, or two return reasons:\newline
  aleatoric\newline
- The user needs to choose a refund method, item option, or other valid
  preference among acceptable alternatives:\newline
  aleatoric\newline
- The user already gave every factual detail, but the next step is to ask for a\newline
  preference or choice from a short list:\newline
  aleatoric\newline
- The user has confirmed the factual record, and the next step is to ask which
  option they prefer:\newline
  aleatoric\newline
- The user says "I don't care", "whatever is cheaper", or "surprise me" and 
  the only unresolved part is the user's choice:\newline
  aleatoric\newline
- The assistant would need to ask the user "yes/no", "which one", or "what
  should I do" because the choice itself is unresolved:\newline
  aleatoric\newline
- The agent already has enough facts to proceed with a tool call or a direct response:\newline
  no\_uncertainty\newline
- The user has already stated the preference and no further question is
  needed:\newline
  no\_uncertainty\newline
- The assistant can directly call a tool or finish the request without asking anything else:\newline
  no\_uncertainty
\par\smallskip
Output format (strict): a single line containing ONLY the label, no\newline
explanation, no punctuation. Example: epistemic
\par\smallskip
[Dialogue history at decision point]\newline
\{history\}
\par\smallskip
[Label]
\end{tcolorbox}

\subsubsection{Ablation Prompt Templates}

The \textsc{LLM Router} ablation replaces goal-conditioned decomposition with
a single three-way prompted decision while keeping the downstream query
generator unchanged.

\begin{tcolorbox}[breakable, colback=gray!20, colframe=gray!50, title={LLM Router}, fontupper=\ttfamily\footnotesize]
You are a routing policy for a tau-bench customer-service agent.
\par\smallskip
Choose exactly one next mode for the agent:
\par\smallskip
- act: the agent should take the next normal tool or final-response action.\newline
- clarify: the agent needs information that only the user can answer, such as
  preference, intent, or an ambiguous user-owned choice.\newline
- verify: the agent needs world-side evidence, policy evidence, computation, or
  a read-only lookup before taking a mutating action or final answer.
\par\smallskip
Prefer act when the next step is clear. Prefer verify over clarify when the
missing information is retrievable from tools, prior tool outputs, policy text,
or deterministic calculation. Prefer clarify only for user-owned ambiguity.
\par\smallskip
Available action/tool names:\newline
\{tools\}
\par\smallskip
Dialogue history:\newline
\{history\}
\par\smallskip
Return strict JSON only:\newline
\{\{"mode": "act" | "clarify" | "verify", "rationale": "short reason"\}\}
\end{tcolorbox}

\end{document}